\documentclass[lettersize,journal]{IEEEtran}
\usepackage{amsmath,amsfonts}
\usepackage{algorithmic}
\usepackage{algorithm}
\usepackage{array}
\usepackage[caption=false,font=normalsize,labelfont=sf,textfont=sf]{subfig}
\usepackage{textcomp}
\usepackage{stfloats}
\usepackage{url}
\usepackage{verbatim}
\usepackage{graphicx}
\usepackage{cite}
\usepackage{amsmath,amssymb,amsfonts}
\usepackage{algorithmic}
\usepackage{graphicx}
\usepackage{textcomp}
\usepackage{makecell}
\usepackage{amsmath,amsfonts}
\usepackage{algorithmic}
\usepackage{algorithm}
\usepackage{array}
\usepackage[caption=false,font=normalsize,labelfont=sf,textfont=sf]{subfig}
\usepackage{textcomp}
\usepackage{stfloats}
\usepackage{url}
\usepackage{verbatim}
\usepackage{graphicx}
\usepackage{cite}
\usepackage{booktabs,multirow}
\usepackage{colortbl}
\usepackage{pifont}
\usepackage{threeparttable}
\definecolor{ccr}{RGB}{0,76,153}  
\usepackage{hyperref}
\hypersetup{
pdfauthor=UnnamedOrange
      hypertex=true,
	colorlinks=true,
	linkcolor=ccr,
	filecolor=blue,      
	urlcolor=ccr,
	citecolor=green,
      anchorcolor=blue,
}
\begin{document}

\title{Dual-Domain CLIP-Assisted Residual Optimization Perception Model for Metal Artifact Reduction}
\author{Xinrui Zhang, Ailong Cai, Shaoyu Wang, Linyuan Wang, Zhizhong Zheng, Lei Li, and Bin Yan 
\thanks{This work is supported by the National Natural Science
Foundation of China (Grant Nos: 62101596 and 62271504).  (Corresponding authors: Lei Li and Bin Yan.) }
\thanks{Xinrui Zhang, Ailong Cai, Linyuan Wang, Zhizhong Zheng, Lei Li,
and Bin Yan is with the Henan Key Laboratory of Imaging and
Intelligent Processing, Information Engineering University, Zhengzhou
450001, China. (e-mail: zxr200043@163.com; cai.ailong@163.com; wanglinyuanwly@163.com; zhengzz81@163.com; leehotline@163.com; ybspace@hotmail.com)
}%
\thanks{Shaoyu Wang is with the Key Laboratory of
Optoelectronic Technology and Systems, Ministry of Education, Chongqing
University, Chongqing 400044, China. (e-mail: wangsy1@cqu.edu.cn)}}

\markboth{}%
{Zhang \MakeLowercase{\textit{et al.}}: A Sample Article Using IEEEtran.cls for IEEE Journals}


\maketitle

\begin{abstract}
Metal artifacts in computed tomography (CT) imaging pose significant challenges to accurate clinical diagnosis. The presence of high-density metallic implants results in artifacts that deteriorate image quality, manifesting in the forms of streaking, blurring, or beam hardening effects, etc. Nowadays, various deep learning-based approaches, particularly generative models, have been proposed for metal artifact reduction (MAR). However, these methods have limited perception ability in the diverse morphologies of different metal implants with artifacts, which may generate spurious anatomical structures and exhibit inferior generalization capability. To address the issues, we leverage visual-language model (VLM) to identify these morphological features and introduce them into a dual-domain CLIP-assisted residual optimization perception model (DuDoCROP) for MAR. Specifically, a dual-domain CLIP (DuDoCLIP) is fine-tuned on the image domain and sinogram domain using contrastive learning to extract semantic descriptions from anatomical structures and metal artifacts. Subsequently, a diffusion model is guided by the embeddings of DuDoCLIP, thereby enabling the dual-domain prior generation. Additionally, we design prompt engineering for more precise image-text descriptions that can enhance the model’s perception capability. Then, a downstream task is devised for the one-step residual optimization and integration of dual-domain priors, while incorporating raw data fidelity. Ultimately, a new perceptual indicator is proposed to validate the model's perception and generation performance. With the assistance of DuDoCLIP, our DuDoCROP exhibits at least 63.7\% higher generalization capability compared to the baseline model. Numerical experiments demonstrate that the proposed method can generate more realistic image structures and outperform other SOTA approaches both qualitatively and quantitatively. 
\end{abstract}

\section{Introduction}
\label{sec:introduction}
\IEEEPARstart{C}{omputed} tomography (CT) imaging has revolutionized the field of medical diagnosis, providing doctors with detailed clinical information\cite{10361833,9770134,10121706}. However, metal implants and other high-density materials embedded in internal organs, bones, and structures can introduce severe artifacts. The reshaped sinogram of collected data is corrupted due to the high attenuation and scattering of X-rays by metal objects\cite{1987Reduction}. This streak-like artifact can affect the quality of CT images, hindering the clinical diagnosis. Hence, how to effectively eliminate metal artifacts of different locations, shapes, and quantities remains a challenge. 


The exploration for metal artifact reduction (MAR) using deep learning (DL)-based methods has recently garnered significant attention. These methods typically involve processing in both image and sinogram domains. However, single-domain knowledge is insufficient for MAR tasks, {\it i.e.}, image-domain methods lack data consistency in handling global artifacts\cite{8331163}, while sinogram-domain methods inevitably introduce severe secondary artifacts\cite{yu2020deep}. To address these issues, joint dual-domain MAR networks have been proposed consecutively. For instance, DuDoNet\cite{Lin2019DuDoNetDD} was proposed as an end-to-end trainable dual-domain network that sequentially refined sinograms and images, respectively. To achieve better disentanglement of artifact distributions in the latent space, the unsupervised generative model performs realistic effects that closely mimic real data distribution, including GAN\cite{xie2024gan}, VAE\cite{togo2024concvae}, Flow-based model\cite{kingma2018glow}, and diffusion model\cite{karageorgos2024denoising}, {\it etc}. Based on the score-based diffusion model, a dual domain method with diffusion priors called DuDoDp\cite{liu2024unsupervised} was proposed and improved its effectiveness for unsupervised MAR. 
Although the diffusion model demonstrates insightful performance in MAR tasks, it remains susceptible to generating spurious structures in the presence of metal artifacts exhibiting diverse morphologies. This limitation hinders its effect in terms of perceptual awareness and generalization capabilities.

\begin{figure}[!t]
\centering
\includegraphics[scale=0.112]{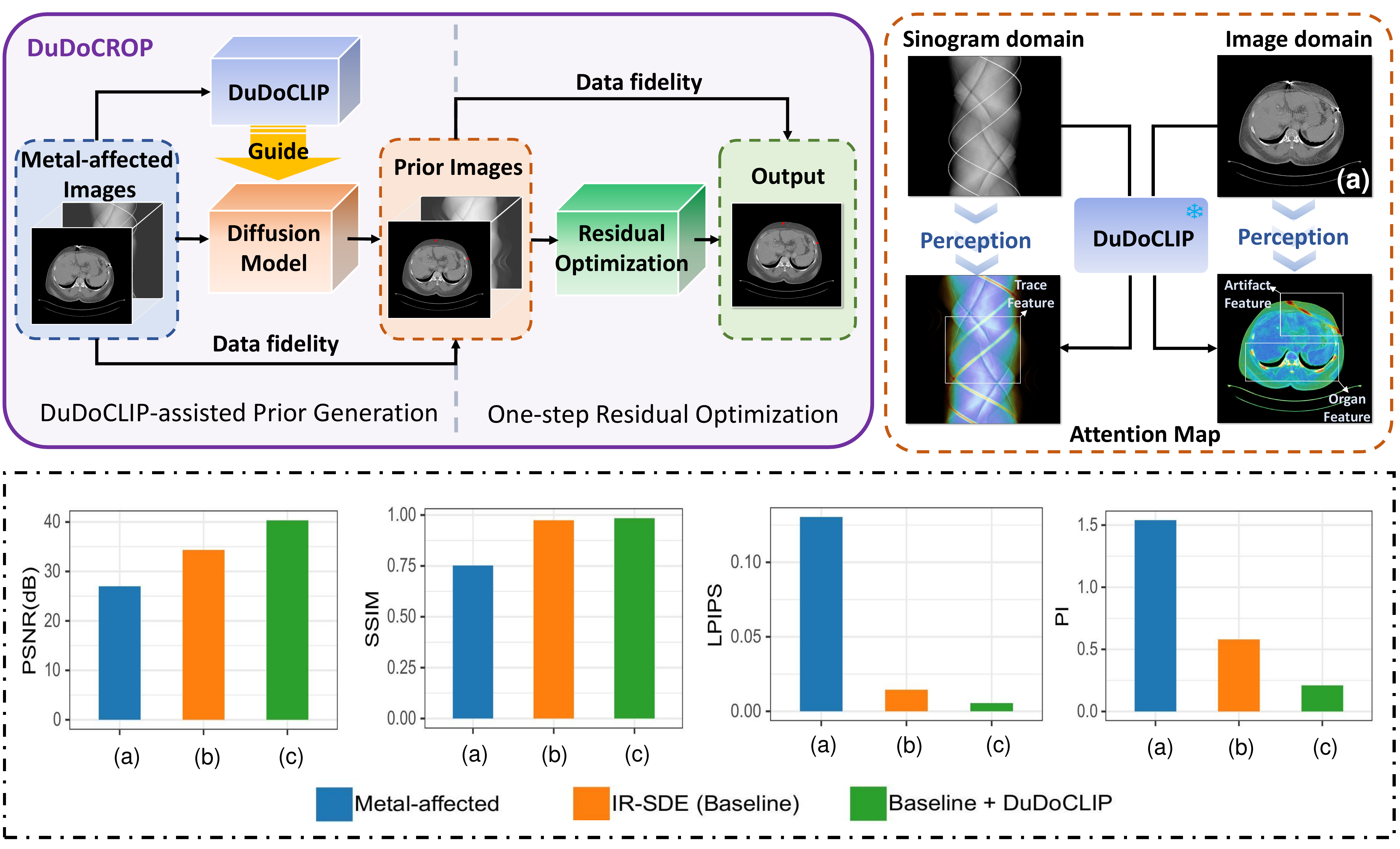}%
\caption{The simplified framework of our DuDoCROP and fundamental principles of the DuDoCLIP model. The metrics "PSNR/SSIM/LPIPS/PI" are used to evaluate the effect of the DuDoCLIP model. Where the average PI=(PI$_s$+PI$_q$)/2 is proposed in section III.D. Comparisons: (a) Metal-affected images (b) IR-SDE (c) IR-SDE w/ DuDoCLIP}
\label{Perception}
\vspace{-1.5em}
\end{figure}

\begin{figure}[!t]
\centering
\includegraphics[scale=0.26]{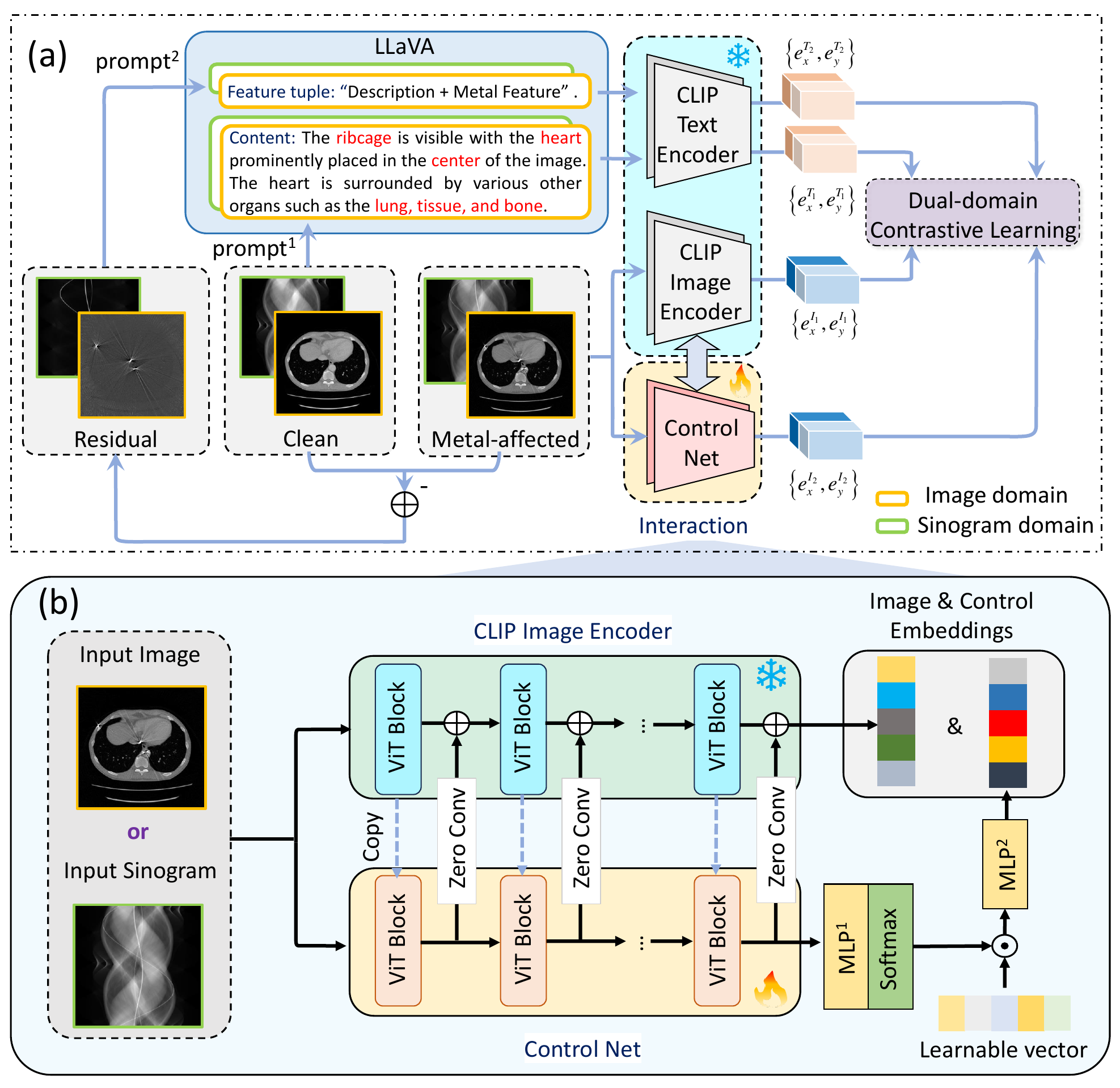}%
\caption{The specific structure of proposed DuDoCLIP. (a) The overall architecture of DuDoCLIP. (b) The interaction of the image encoder and control net. The control net is the copy of the CLIP image encoder which is the cascaded visual transformer (ViT) blocks\cite{dosovitskiy2020image} architecture. The zero convolution layer allows for the introduction of conditional control while maintaining model stability, enabling fine adjustments to the generated embeddings.}
\label{The specific structure}
\vspace{-1.5em}
\end{figure}

Currently, the foundation models (FMs) demonstrate astonishing capabilities in text generation and knowledge reasoning, and have spurred a series of visual-language interactive generative models\cite{tan2022dr}. In medical imaging, the image-text-based visual-language model (VLM) has been applied as a paradigm to various downstream tasks. Common VLMs that assist in medical diagnosis include ViLBERT\cite{lu2019vilbert}, LLaVA\cite{li2024llava}, ChatGPT\cite{wu2023visual}, and others. The existing application for VLM is to generate multi-modal diagnostic and treatment reports. For example, Medical-VLBERT\cite{liu2021medical} was proposed to generate medical reports automatically for the diagnosis of COVID-19. 

ChatRadio-Valuer\cite{zhong2023chatradio} was trained on the radiology report using supervised fine-tuning on VLMs and then adapted to multiple disease diagnosis tasks. Especially, it has outperformed GPT-3.5-Turbo and GPT-4 et al. To integrate VLM with low-level imaging issues, a low-dose CT denoising method (LEDA)\cite{chen2024low} was proposed to introduce text tokens into the quantizer of GAN. Therefore, it has been proven that foundation models have insightful generalization ability in multi-task and multi-modal imaging tasks.

To tackle the challenges of limited model generalization and imaging efficacy for diverse metal artifact morphologies in current MAR-specific methods, we are motivated to introduce the VLM-based foundation model structured akin to CLIP\cite{hafner2021clip}, thereby identifying these morphological features as semantic information, which concurrently assists in MAR within both the image and sinogram domains. Specifically, we propose a dual-domain CLIP (DuDoCLIP) model, capable of distinguishing diverse organ structures and artifacts of different metal implants, thereby embedding high-level semantic information of VLMs to guide and optimize the diffusion model. As shown in Fig. \ref{Perception}, the DuDoCLIP is introduced in the first stage of our two-stage MAR framework, which endows the diffusion model with a "perceptual ability", enabling it to adaptively perceive metal-affected images through an attention mechanism tailored to the different metal-affected regions in dual domains. This perceptual model contains VLM priors, fostering generalization and minimizing spurious structures in the restored CT images, thereby improving the final imaging effects according to various metrics. In addition, to further reduce the spurious structures caused by generative models and enhance data fidelity, we design a downstream task as the second stage for residual optimization, which refines and integrates the dual-domain prior images. Hence, the entire framework constitutes our DuDoCLIP-assisted residual optimization perception model, named \textbf{DuDoCROP}.

At the DuDoCLIP-assisted prior generation (DAPG) stage, as shown in Fig. \ref{The specific structure}, the DuDoCLIP model is fine-tuned on the paired image-domain and sinogram-domain datasets consisting of clean images and their corresponding metal-affected counterparts, leveraging pre-trained CLIP models. We utilize LLaVA to generate text descriptions for clean images and metal implants with artifacts. Through prompt engineering, LLaVA can characterize anatomical structures in clean CTs and extract metal artifact features. The pre-trained CLIP encoders generate embeddings, while a control net, trained via contrastive learning, controls the image encoder to produce high-quality embeddings from metal-corrupted images. 
 The embeddings derived from the image encoder and the control net are used to guide a dual-domain image restoration stochastic differential equation (IR-SDE) diffusion model, facilitated by a cross-attention and integration mechanism. Then, we can obtain the dual-domain prior images. 

At the one-step residual optimization (OSRO) stage, shown in Fig. \ref{The overall architecture}(a), the sinogram-domain prior image corporates the raw data fidelity, forming a residual with the image-domain prior. This residual is then refined through a residual-domain IR-SDE model, further enhancing the image domain prior and reducing the secondary artifacts.  Ultimately, the dual-domain images are integrated to obtain the final output.

Our contributions can be summarized as follows:

\begin{itemize}
    \item 
    We primarily focus on the MAR-specific task leveraging the VLM foundation model and propose a dual-domain CLIP-assisted residual optimization perception model (DuDoCROP). The DuDoCLIP in our model is fine-tuned on the image domain and sinogram domain using contrastive learning to extract semantic descriptions from anatomical structures and metal artifacts, thereby guiding the IR-SDE to generalize on diverse morphologies of different metal implants with artifacts. 
    \item 
    We meticulously design prompt engineering for DuDoCLIP to further enhance the precision of clean image content and metal artifact feature representations, and conclusively verify that the incorporation of more precise image-text descriptions can enhance the model's perception capability.
    \item 
    We derive a downstream task for one-step residual optimization and integration of dual-domain priors, incorporating the raw data fidelity in the sinogram domain. This strategy can effectively reduce secondary artifacts in the sinogram-domain prior and facilitate the restoration of more authentic anatomical structures.
    \item 
    We propose a new perceptual indicator to validate the model's perception and generalization capabilities on diverse metal morphologies of different sizes and quantities. Experiments are conducted on diverse anatomical slices (e.g., body, head, and clinical data), exhibiting superior performance in both visual effects and 
 numerical indicators than other state-of-the-art methods. 
\end{itemize}
\section{Related work}
\subsection{Classic MAR Methods}
When X-rays pass through metal objects, low-energy rays are absorbed in large quantities, manifesting as bright artifacts around the metal in the reconstructed image. Many studies have focused primarily on repairing metal projection data to suppress metal artifacts. Classic methods include linear interpolation (LI)\cite{1987Reduction}, which involves repairing the missing portions of metal sinogram by using linear interpolation based on adjacent no-metal sinogram, as well as polynomial interpolation, and modified linear interpolation\cite{gu2006method}, {\it etc}. The interpolation repair method is straightforward in operation, but it is prone to introducing secondary artifacts. Subsequently, iterative MAR methods (IMAR) achieve good suppression effects with constrained total variation (TV) minimization model\cite{zhang2016iterative}. The metal artifacts can be suppressed by considering the metal implants as priors and continuously iterating and updating during the reconstruction. Especially when combining spectral CT with the IMAR method, it can reduce the artifacts with high-energy bins and improve contrast details with low-energy bins\cite{winklhofer2018combining}. Other MAR approaches have also received much attention, such as Normalized MAR (NMAR)\cite{meyer2010normalized}, Frequency-split NMAR (FSNMAR)\cite{meyer2012frequency}, and segmentation-based MAR\cite{lyu2023pds}, {\it etc}.
\begin{figure*}[!t]
\centering
\includegraphics[scale=0.35]{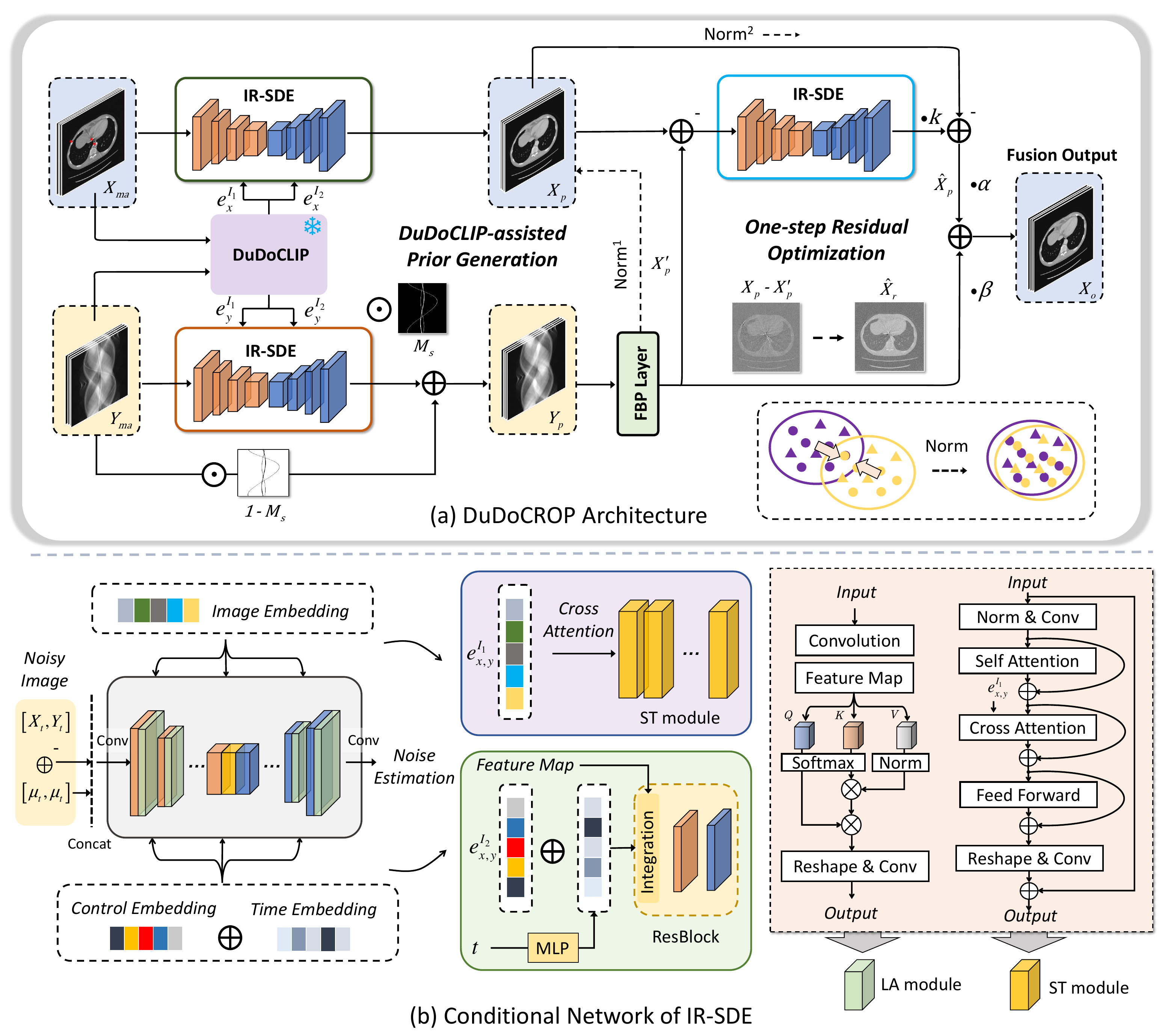}%
\vspace{-1em}
\caption{The overall architecture of proposed DuDoCROP. (a) The inputs of DuDoCROP are paired metal-affected images $X_\text{ma}$ and sinograms $Y_\text{ma}$. At the DuDoCLIP-assisted prior generation (DAPG) stage, the dual-domain IR-SDEs are guided by embeddings $e_{x}^{I_{1,2}}$ and $e_{y}^{I_{1,2}}$, respectively. The prior images $X_p$ and $Y_p$ are generated with the assistance of DuDoCLIP. At the one-step residual optimization (OSRO) stage, the data fidelity from the raw sinogram $Y_\text{ma}$ is introduced using Eq. (\ref{deqn_ex13a}). The Norm$^1$ is to normalize $X_p$ using $X'_p$, and Norm$^2$ is to normalize $\hat{X}_p$ using $X_p$.  With normalization, the distance between the two distributions is reduced. The residual $X_p-X'_p$ is refined as $\hat{X}_r$ through IR-SDE and utilized to update image-domain prior as $\hat{X}_p$ using Eq. (\ref{deqn_ex16a}). Finally, the two priors are fused with factors for the final output. (b) The image and control embeddings are introduced into a conditional network of IR-SDE through integration and cross-attention (explained in the second part of section III.B). 
}
\label{The overall architecture}
\vspace{-1.5em}
\end{figure*}
\vspace{-1em}
\subsection{DL-based MAR Methods}
DL-based methods provide new insights for MAR with their powerful feature extraction and nonlinear mapping capabilities. One of the earliest and most widely used deep learning methods for MAR is based on Convolutional Neural Networks (CNNs). Zhang et al. first applied CNNs to metal artifact suppression, proposing an end-to-end MAR network architecture with a simple encoder and decoder\cite{8331163}. At present, generative models demonstrate powerful performance and have been widely applied in MAR. Artifact Disentanglement Network which is the so-called ADN\cite{8788607} can disentangle the metal artifacts from CT images in the latent space. Unsupervised CycleGAN\cite{lee2021unsupervised} is optimized by adding the convolutional block attention module, improving metal artifact reduction that preserves the detailed texture. Another type of generative diffusion model is also currently a focus. Song et al. proposed a score-based model \cite{song2020score} adopting reverse-time stochastic differential equations (SDE) to sample the data distribution from a known prior distribution. Other state-of-the-art methods include DDIM\cite{song2021denoising}, DPS\cite{dou2024diffusion}, and some novel methods specifically designed for MAR, such as DEVOTED-Net\cite{tong2023data}, DuDoDp-MAR\cite{liu2024unsupervised}, and 3D CBCT-MAR\cite{choi2024dual}, {\it etc}. In addition to the single-domain MAR method, the multi-domain cascaded network can fully utilize the information from different domains. Such methods including DuDoNet\cite{Lin2019DuDoNetDD}, InDuDoNet\cite{inbook}, and DuDoDR-Net\cite{ZHOU2022102289} combine information from the image domain and the sinogram domain, while Quad-Net\cite{li2024quad} innovatively introduces Fourier domain information. Additionally, multi-domain networks with a parallel structure can further reduce secondary artifacts, like IDOL-Net\cite{9765584}.
\vspace{-1em}
\subsection{VLM-Assisted Image Generation}
With the arrival of another revolution in artificial intelligence (AI), the foundation models demonstrate great potential in cross-domain interactions, leveraging big data training and powerful model architectures. The VLM \cite{hafner2021clip,long2023fine} has been regarded as a representative multimodal foundation model, employing visual and linguistic encoder-decoders with aligned character representations of image and text modalities. This allows embedding high-level semantic information into low-level tasks with pre-trained vision-language models, such as image inpainting, noise suppressing, and super-resolution, {\it etc}. For example, in the computer vision field, a state-of-the-art generative model called Stable Diffusion (SD) \cite{rombach2022high} can couple text embeddings with the U-Net in the form of cross-attention, which has been studied on a wide range of tasks including text-to-image generation, image-to-image translation, image restoration, and more. In particular, SUPIR\cite{yu2024scaling} was proposed to reuse an adaptive multi-modal VLM to generate specific text prompts that guided the super-resolution and denoising tasks of the diffusion model. DA-CLIP\cite{luo2023controlling} was proposed as a multi-task unified framework using pre-trained vision-language models for image-text contrastive learning on large-scale datasets with multiple types of degradations, including "blurry", "hazy", "low-light", and "noisy", {\it etc}. The image-text pairs of DA-CLIP can integrate different degradation types of knowledge from VLM into a general network encoder, thereby enhancing the awareness capability of the classical CLIP.
\section{Methodology}
\subsection{Overview}
\subsubsection{Modeling of MAR Problem}We firstly denote $X\in \mathbb{R}^{H\times W}$ and $Y\in \mathbb{R}^{U\times V}$ as 2-D CT image and corresponding sinogram, respectively. $H,W$ and $U,V$ are corresponding height and weight. The sinogram $ {Y}$ can be obtained by forward projection, denoted as $ Y=FP(X)$, and CT images ${X}$ can be reconstructed using filtered back-projection (FBP), denoted as $ X=FBP(Y)$. Metal artifact reduction (MAR) aims to recover clean CT images $ {X_\text{ref}}$ from metal-affected images $ {X}_{\text{ma}}$ or corrupted sinograms $ {Y}_{\text{ma}}$ to obtain real sinograms $ {Y_\text{ref}}$.  A binary metal mask $M_s$ can be segmented using thresholding from the metal-affected regions (set to 1) in ${Y}_{\text{ma}}$, and the other regions of $M_s$ are set to 0. Thus, the relationship of sinograms $Y_{\text{ma}}$ and $Y_{\text{ref}}$ can be written as:
\begin{equation}
\label{deqn_ex1a_add}
Y_\text{ma}\odot \left( 1-M_s \right)=Y_\text{ref}\odot \left( 1-M_s \right)
\end{equation}
 Where $\odot$ is the Hadamard product. Thus, the non-destructive region of the raw data ${Y}_{\text{ma}}$ contains data fidelity from $Y_\text{ref}$.
\subsubsection{Architecture of DuDoCROP} The overall workflow of our DuDoCROP is illustrated in Fig. \ref{The overall architecture}(a). The DuDoCROP is composed of a DuDoCLIP-assisted prior generation (DAPG) stage and a one-step residual optimization (OSRO) stage. The DAPG stage utilizes pre-trained DuDoCLIP to perform fine-grained perception on different anatomical structures with metal artifacts and generate paired image embeddings and control embeddings.  The dual-domain IR-SDE  diffusion model is guided by these embeddings to produce high-quality priors in both image domain and sinogram domain. 
The OSRO stage as a downstream task in DuDoCROP can deeply explore residual information to reduce the sinogram-domain secondary artifacts and fully integrate the dual-domain prior information, and the data fidelity from the raw sinogram is introduced to the image-domain priors to generate more realistic anatomical structures.

Based on this, our DuDoCROP adopts an ensemble learning strategy that requires three-stage training including DuDoCLIP fine-tuning, dual-domain IR-SDE learning, and residual refinement.
\vspace{-1em}
\subsection{DAPG Stage with DuDoCLIP Embedding Guidance}
\subsubsection{DuDoCLIP Architecture with Visual-Launage Foundation Model}
DA-CLIP\cite{luo2023controlling} is a visual-language foundation model that efficiently handles multiple image restoration tasks, such as image denoising, deblurring, inpainting and Low-light image enhancement, {\it etc.} Although DA-CLIP has been performed on various restoration tasks of natural images, this model has not yet been applied to MAR-specific tasks in CT imaging. Hence, inspired by DA-CLIP, we design the DuDoCLIP to enable a more fine-grained perception on different metal-affected anatomical structures with artifacts. As shown in Fig. \ref{The specific structure}(a), we adopt the LLaVA foundation model to generate text descriptions for both clean images and residual images containing metal implants with artifacts. With meticulous prompt engineering, LLaVA can more precisely characterize organs such as the heart, tissues, and other anatomical structures in clean CT images as content descriptions, while extracting metal and artifact features including size, location, quantity, and artifact descriptions within structured tuples. Subsequently, the pre-trained CLIP's image and text encoders are engaged to generate image and text embeddings. 

\begin{figure*}[!t]
\centering
\includegraphics[scale=0.28]{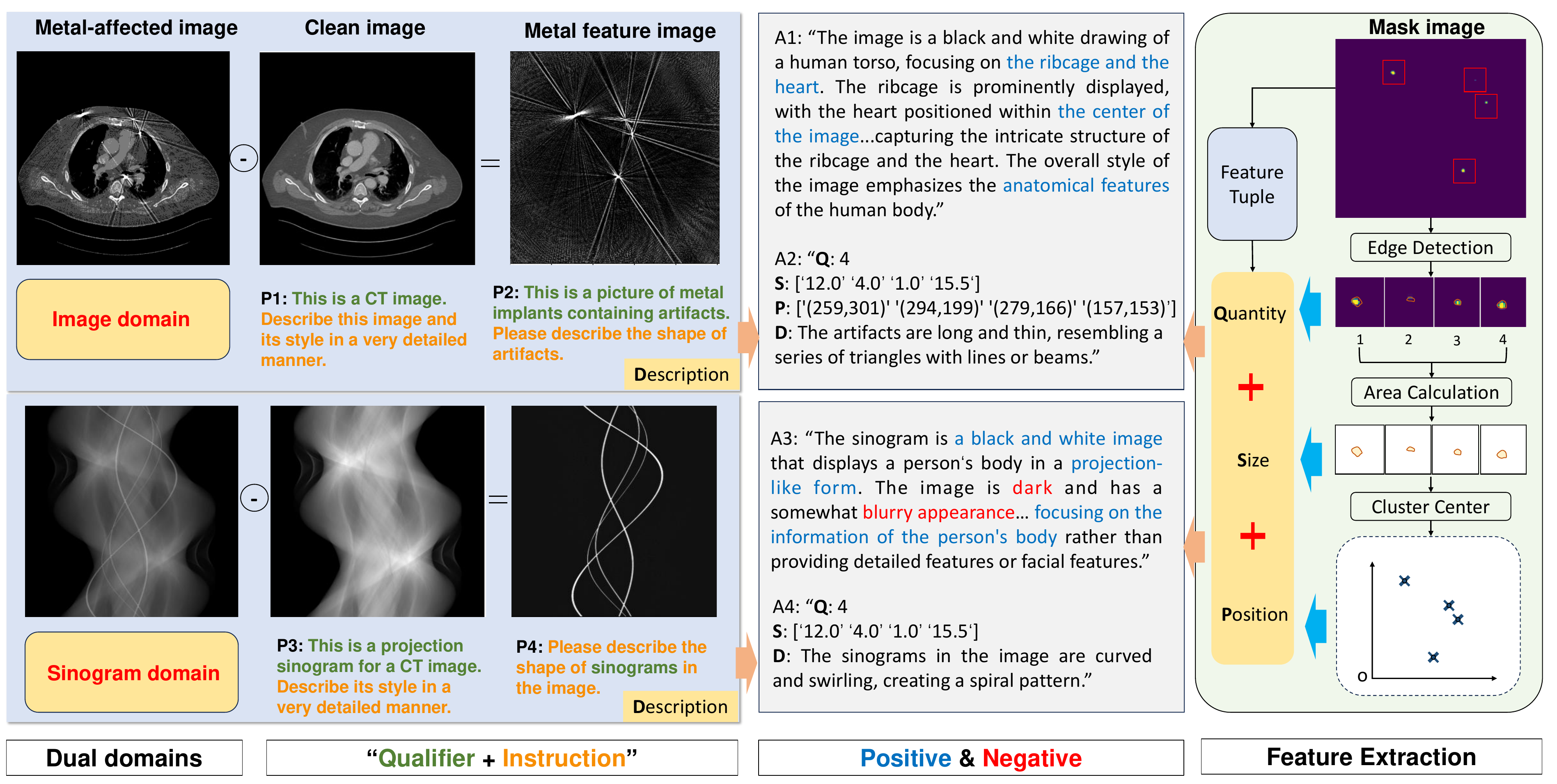}%
\caption{The prompt engineering of DuDoCLIP. We adopt the prompt pattern of "Qualifier + Instruction" to generate the expected text description. A1-A4 are descriptions of prompts P1-P4, respectively. The accuracy of the prompt determines the proportion of positive and negative(P/N) text description words.}
\label{The prompt engineering}
\vspace{-1em}
\end{figure*}

The control net, interacting with the image encoder, is trained by contrastive learning, thereby controlling the image encoder to generate high-quality embeddings of clean images from the metal-corrupted images. Furthermore, the architecture of the control net is the copy of image encoder which is a cascaded ViT, as shown in Fig. \ref{The specific structure}(b), they achieve information interaction through zero convolution\cite{zhang2023adding}.  
\begin{table}
\centering
\caption{THE PREPARATION OF DIFFERENT DATASETS }
\arrayrulecolor{black}
\arrayrulecolor{black}
\begin{tabular}{ccc}
\toprule[0.3mm]
\multirow{1}{*}{Dataset}
&\multirow{1}{*}{AAPM(CT-MAR)}
&\multirow{1}{*}{CTPelvic1K}
\\
\cmidrule(l){1-3}
\multirow{2}{*}{Training
  Set}
  &\multirow{2}{*}{2475
  body, 528 head}                      &\multirow{2}{*}{None}                            \\ \\
  \cmidrule(l){1-3}
\multirow{2}{*}{Testing
  Set}                                     & \multirow{2}{*}{864 body\textsuperscript{1},  642 body\textsuperscript{2}, 136 head }& \multirow{2}{*}{ 200
  body}                      \\ \\ 
\arrayrulecolor{black}
\bottomrule[0.3mm]
\end{tabular}
\arrayrulecolor{black}
\label{Table1}
\vspace{-2.5em}
\end{table} In summary, the prompt engineering and contrastive learning mentioned are significant operations performed in DuDoCLIP:

\textbf{Prompt Engineering:}
Specifically, the clean image feature is translated as content description including organ type, location, color, {\it etc.} The residual images denoted as $ {X}_{\text{r}}$ and $ {Y}_{\text{r}}$ contain distinctive metallic characteristics, which can be obtained by $ {X}_{\text{r}}={X}_{\text{ma}}-{X_{ref}}$ and $ {Y}_{\text{r}}={Y}_{\text{ma}}-{Y_{ref}}$. The description of residual images reflects metallic morphology and artifact features. Under meticulously designed prompts, this feature is presented in the form of a structured tuple that includes the size, quantity, and position of metallic implants, as well as the description of artifacts. Therefore, the accuracy of text description significantly relies on prompt design. As can be seen in Fig. \ref{The prompt engineering}, the prompt words consist of "Qualifier + Instruction" to improve the accuracy of content description (A1 \& A3) and tuple description (A2 \& A4). Additionally, to accurately describe the metal implants and corresponding sinogram features in the residual images, we construct a feature tuple composed of \textbf{Q}uantity, \textbf{S}ize, \textbf{P}osition and \textbf{D}escription (QSPD). Where the description of the tuple is obtained from the LLaVA foundation model, while features such as quantity, size, and position can be easily extracted from the mask image.

\textbf{Contrastive Learning:}
Then, the clean image contents and metal feature tuples are fed to the frozen CLIP text encoder to obtain embeddings $
\left\{ e_{x}^{T_{1}},e_{y}^{T_{1}} \right\} $ and $\left\{e_{x}^{T_{2}},e_{y}^{T_{2}} \right\} $, respectively. Where $
e_{x}^{(\cdot)}  $ and $
e_{y}^{(\cdot)} $ are denoted as image-domain and sinogram-domain text embeddings. The CLIP image encoder is also frozen to transfer the metal-affected images and sinograms as image embeddings $
\left\{ e_{x}^{I_{1}},e_{y}^{I_{1}} \right\} 
$.  Hence, only the control net is fine-tuned by interaction with the image encoder and produces control embeddings $
\left\{ e_{x}^{I_{2}},e_{y}^{I_{2}} \right\} 
$ for dual-domain contrastive learning. This enables metal-affected images and clean image descriptions to be paired in the embedding space, while metal-affected images and metal feature descriptions are also paired. We assume the number of paired embeddings in the single domain is $N$. The $i$th positive  sample and $j$th negative sample are denoted as vectors $\boldsymbol{a}_{i}, i=1,\dots,N$ and $\boldsymbol{b}_{j}, j=1,\dots,N$, respectively. The distance between each paired sample is defined as
\begin{equation}
\label{deqn_ex1a}
\mathcal{D}_{\text{con\,\,}}\left( \boldsymbol{a,b} \right) =-\log \left( \frac{\exp \left( \boldsymbol{a}_{i}^{\top}\boldsymbol{b}_i/\tau \right)}{\sum_{j=1}^N{\exp}\left( \boldsymbol{a}_{i}^{\top}\boldsymbol{b}_j/\tau \right)} \right), 
\end{equation}
Where $\tau$ is a learnable temperature parameter that is used to control the weights between different sample pairs in the contrastive loss $\mathcal{L}_{\text{con}}$. Based on this, the dual-domain contrastive loss can be written as 
\begin{equation}
\label{deqn_ex2a}
\!\!\!\!\mathcal{L}_{\text{con\,\,}}\!\!\left( \varphi \right)\!\!=\!\!\frac{1}{N}\sum_{i=1}^N{\sum_{s=1}^2{\mathcal{D}_{\text{con\,\,}}\!\!\left( e_{x}^{I_s},e_{x}^{T_s};\varphi \right) +\mathcal{D}_{\text{con\,\,}}\!\!\left( e_{y}^{I_s},e_{y}^{T_s};\varphi \right)}}.
\end{equation}
\begin{figure}[!t]
\centering
\includegraphics[scale=0.3]{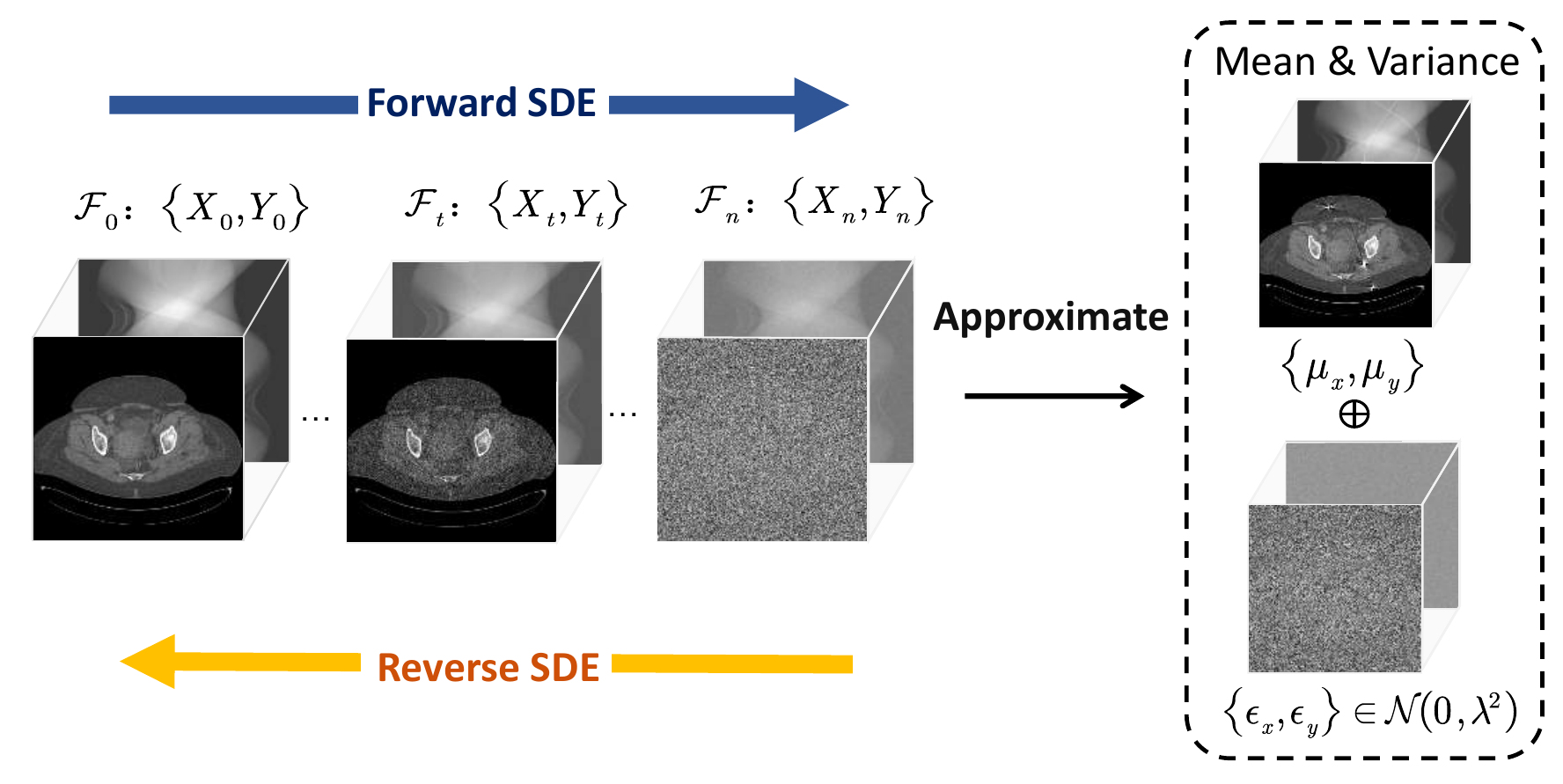}%
\vspace{-1em}
\caption{The IR-SDE diffusion model contains forward SDE and reverse SDE. Each step in the reverse process can be regarded as a prediction of the mean and variance}
\label{The IR-SDE diffusion}
\vspace{-1.5em}
\end{figure}
In this way, the trained control network and image encoder can generate high-quality clean text descriptions and metal feature descriptions from the dual-domain metal-affected images, thereby providing references for the diffusion model guided by conditional mechanisms.
\subsubsection{ Conditionally Dual-domain Embedding Guided IR-SDE Diffusion Model}
The IR-SDE diffusion model\cite{luo2023image} is the key module in DuDoCROP, which is applied in image-domain artifact reduction, sinogram-domain metal trace measurement restoration, and residual-domain refinement, as shown in Fig. \ref{The overall architecture}. The IR-SDE consists of a forward
diffusion process and a reverse denoising process, as shown in Fig. \ref{The IR-SDE diffusion}. For a continuous time variable of $t>0$, we consider a dual-domain diffusion process that follows the marginal distribution at time $t$: ${X_{t}, Y_{t}} \sim {p_{t}(X),p_{t}(Y)}$. To simplify the expression, we denote a diffusion process ${\mathcal{F}_{t}}\in\{X_{t}, Y_{t}\}$ as a symbol state of the image domain or sinogram domain. 

\textbf{IR-SDE Mathematical Model:}
  In this paper, We construct a dual-domain form of the analytical forward SDE score function denoted as follows:
\begin{equation}
\label{deqn_ex3a}
d{\mathcal{F}}={\gamma }_{t}\left( {\mu}-{\mathcal{F}} \right) dt+{\sigma }_{t}d\omega, 
\end{equation}
Where ${\mu}$ refers to the mean of the state. ${\gamma}_{t}$ and ${\sigma} _{t}$ are scalar-valued time-dependent parameters.  $\omega$ is the standard Wiener process. According to a reverse-time representation of the SDE verified by Anderson (1982)\cite{anderson1982reverse}, Eq.(\ref{deqn_ex3a}) can be derived as the reverse form of IR-SDE: 
\begin{equation}
\label{deqn_ex4a}	d{\mathcal{F}}=\left[ {\gamma} _{t}\left({\mu}-{\mathcal{F}} \right) -{\sigma} _{t}^{_2}\nabla_{{\mathcal{F}}}\log p_t\left({\mathcal{F}} \right) \right] dt+{\sigma}_{t}d\widehat{\omega }, 
\end{equation}
\begin{figure}[!t]
\centering
\includegraphics[scale=0.18]{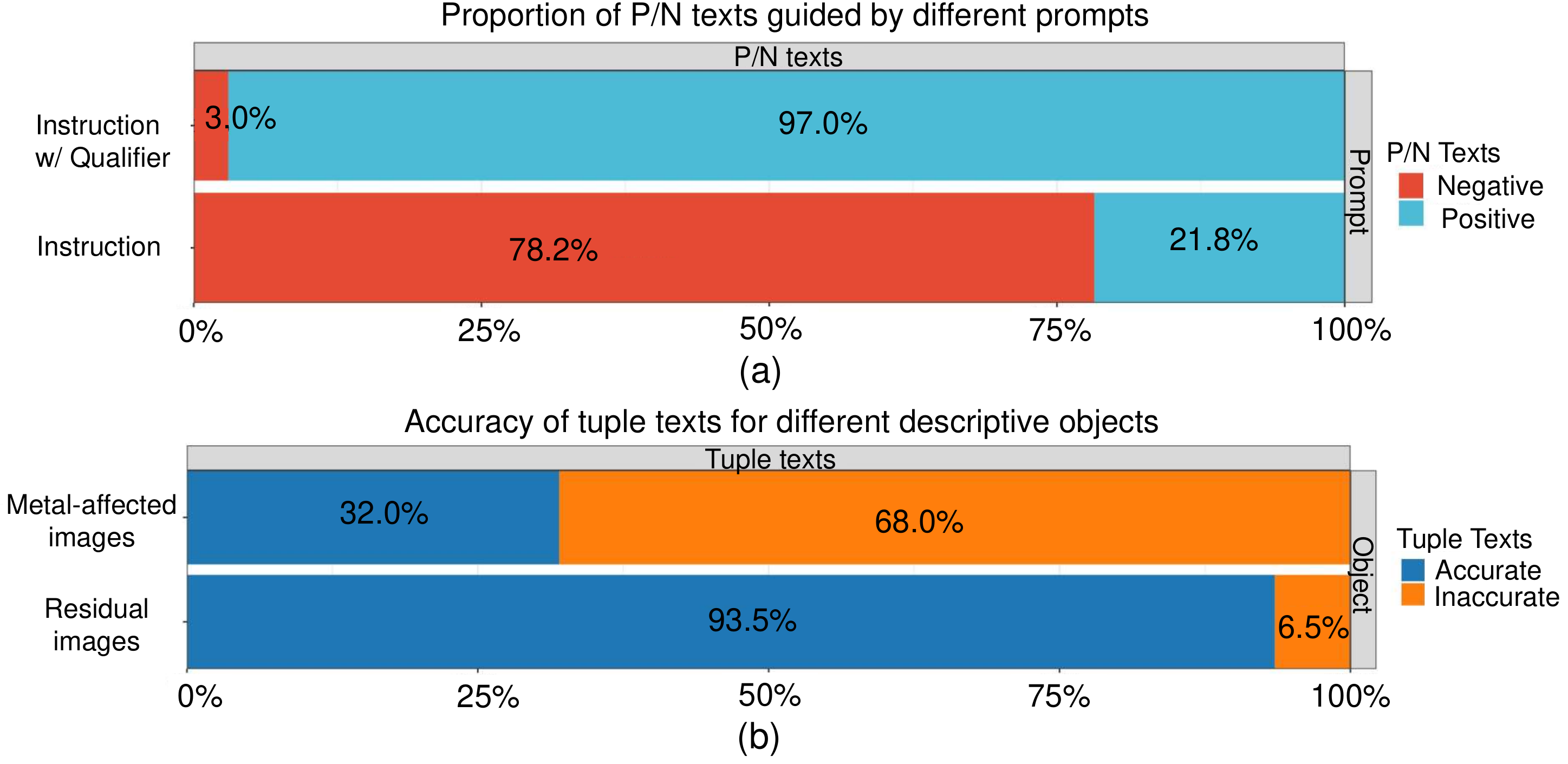}%
\caption{Statistical results showing the impact of different prompts and described objects on the proportion of positive and negative (P/N) text. (a) is the proportion of P/N texts guided by prompts with or without qualifiers. (b) is the accuracy of tuple text in the case of directly describing Metal-affected images or describing metallic characteristics. }
\label{Statistical results showing}
\vspace{-1em}
\end{figure}
Where $\nabla_{{\mathcal{F}}}\log p_t\left({\mathcal{F}} \right)$ is an unknown real score. In IR-SDE, Eq.(\ref{deqn_ex3a}) has a closed-form solution and a stationary variance ${\lambda}$ is defined as ${\lambda}^{2} = {\sigma}_t^{2}/2{\gamma}_{t}$. After derivation (more details can be found in the supplementary material), IR-SDE defines an explicit expression for the mean $\boldsymbol{m}_{t}$ and the variance $\boldsymbol{v}_{t}$ of the distribution of $p_{t}({\mathcal{F}})$ at each time-step:
\begin{equation}
\label{deqn_ex5a}
\begin{aligned}
	{m}_t:&=\ {\mu}+\left( {\mathcal{F}}_0-{\mu} \right) \text{e}^{-\bar{{\gamma}}_t},\\	{v}_t:&={\lambda} ^2\left( 1-\text{e}^{-2\bar{{\gamma}}_t} \right).\\
\end{aligned}
\end{equation}
\begin{figure}[!t]
\centering
\includegraphics[scale=0.14]{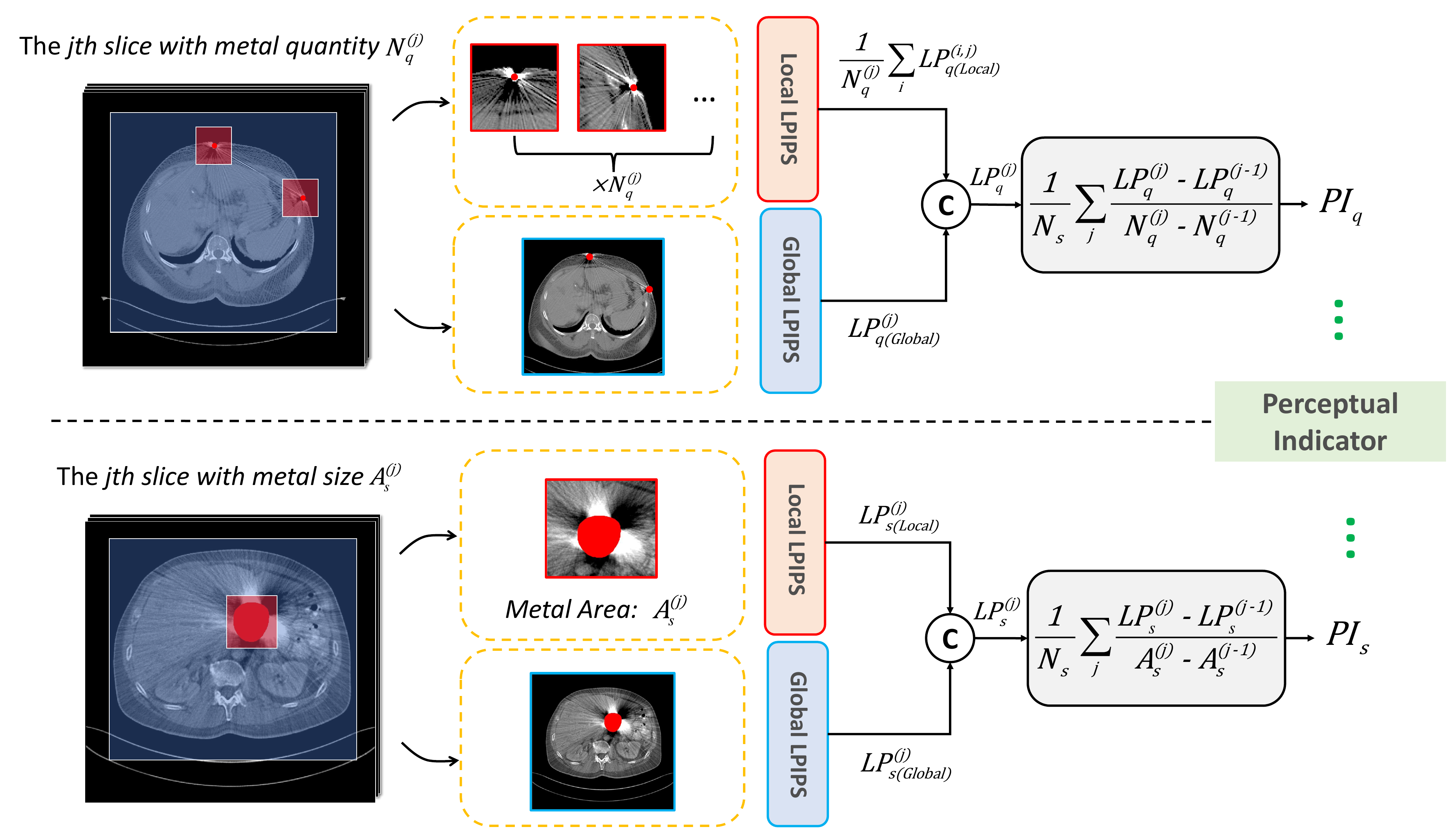}%
\caption{The designing pipeline of proposed perceptual indicators of PI$_q$ and PI$_s$ for metal quantity (1st row) and metal size (2nd row), respectively.}
\label{The designing pipeline}
\vspace{-1em}
\end{figure}
Where $\bar{{\gamma}}_t=\int_0^t{\boldsymbol{\gamma }_z}dz$, ${\mathcal{F}}_0$ is the clean image at $t=0$, and $\mu$ is regarded as the metal-affected counterpart of ${\mathcal{F}}_0$. With $t \to \infty$, ${m}_t$ and ${v}_t$ will converge to $\mu$ and ${\lambda} ^2$. Under the condition that the distribution of clean images is known, the score can be rewritten as $\nabla_{{\mathcal{F}}} \log p_{t}({\mathcal{F}} \mid {\mathcal{F}}_{0})$.In the forward process,  ${\mathcal{F}}_t\sim \mathcal{N}\left( {m}_t,{v}_t \right) $, therefore, ${\mathcal{F}}_{t}$ can be reparameterized as ${m}_{t}+\sqrt{{v}_{t}}{\epsilon}_{t}$, ${\epsilon }_t\sim \mathcal{N}\left( 0,I \right)$. Hence, the precise form of score can be derived  as $\nabla_{{\mathcal{F}}} \log p_{t}({\mathcal{F}} \mid {\mathcal{F}}_{0})=-\frac{{\epsilon }_t}{{v}_{t}}.$ Then, we can construct a score-based network ${\varepsilon }_{\varTheta}({\mathcal{F}}_t, {\mu}, t)$ to predict the noise ${\epsilon }_t$ and thereby fit the ideal score. $\varTheta$ is the network parameter. This is consistent with the idea of score matching in DDPM\cite{ho2020denoising}. Finally, we adopt discrete SDE form and divide $t$ into timestamps $t_i, i=1,2,\dots,n$. A Maximum Likelihood Learning method is applied to solve the reverse IR-SDE which  minimizes a negative log-likelihood:
\begin{equation}
\label{deqn_ex8a}
{\mathcal{F}}_{t_{i-1}}^{*}=\arg \min _{{\mathcal{F}}_{t_{i-1}}}\left[-\log p\left({\mathcal{F}}_{t_{i-1}} \mid {\mathcal{F}}_{t_i}, {\mathcal{F}}_{0}\right)\right],
\end{equation}
Where ${\mathcal{F}}_{t_{i-1}}^{*}$ is the optimal solution at timestamp $t_{i-1}$. Let ${{\gamma'}}_{t_{i}}=\int_{{t_{i-1}}}^{t_{i}}{{\gamma }_z}dz$, and we can obtain the standard solution of ${\mathcal{F}}_{t_{i-1}}^{*}$:
\begin{equation}
\label{deqn_ex9a}
\begin{aligned}
{\mathcal{F}}_{t_{i-1}}^{*}=\frac{1-\text{e}^{-2\overline{{{\gamma} }}_{t_{i-1}}}}{1-\text{e}^{-2\overline{{\gamma }}_{t_i}}}\text{e}^{-{\gamma '}_{t_i}}\left( {\mathcal{F}}_{t_i}-{\mu} \right)\\
	+\frac{1-\text{e}^{-2{\gamma '}_{t_i}}}{1-\text{e}^{-2\overline{{\gamma }}_{t_i}}}\text{e}^{-\overline{{\gamma }}_{t_{i-1}}}\left( {\mathcal{F}}_0-{\mu} \right) +{\mu},
\end{aligned}
\end{equation}
\begin{figure}[!t]
\centering
\includegraphics[scale=0.2]{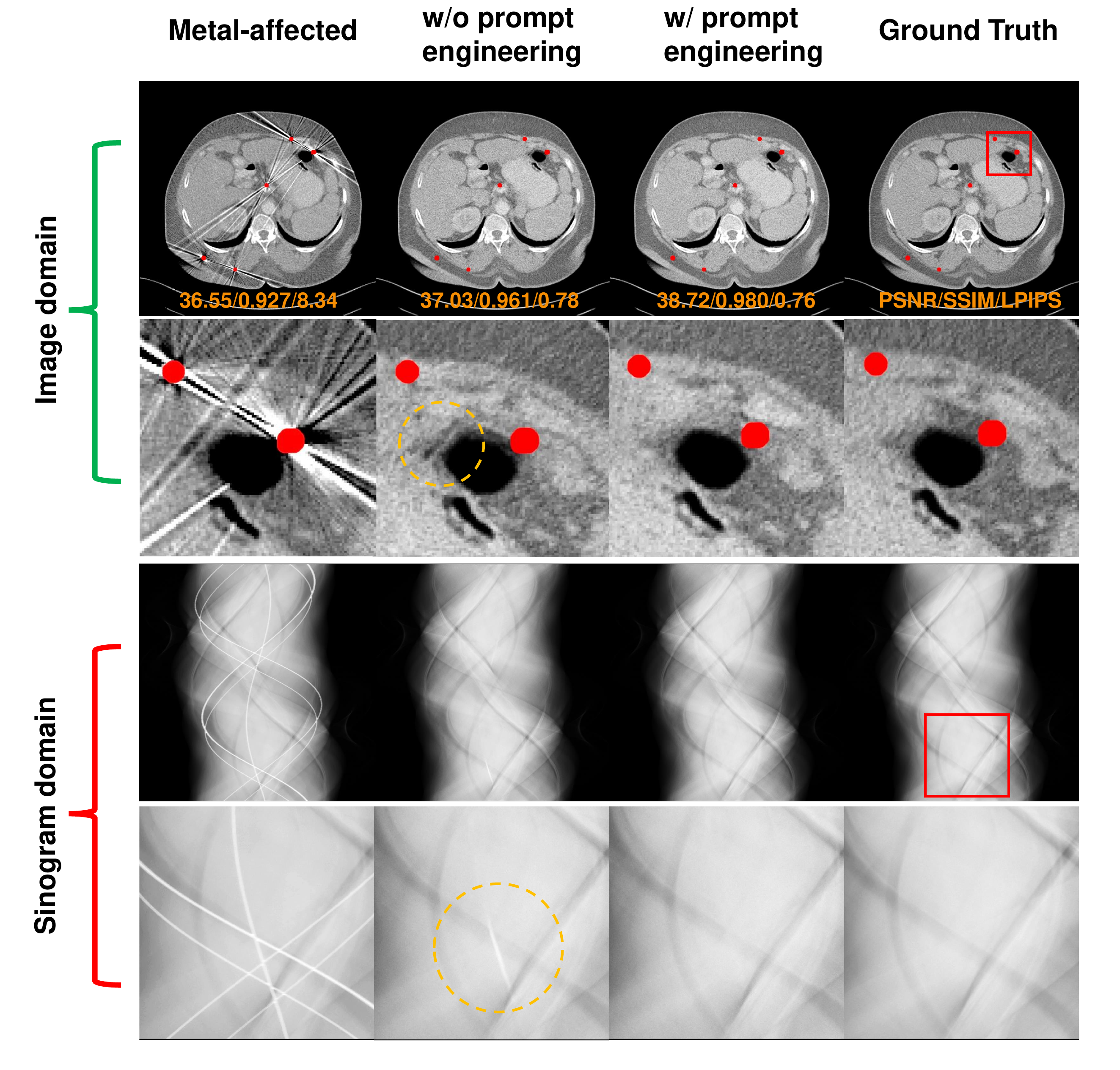}%
\vspace{-1em}
\caption{ The ablation experiment on prompt engineering resulting in the visual effects in dual domains. Quantitative indicators PSNR, SSIM and (LPIPS*100) are applied. Where the  w/o prompt designing is employing a direct description of the image. The display window is [-175, 275] HU.}
\label{The ablation experiment}
\vspace{-1em}
\end{figure}
The actual solution of reverse SDE can introduce score-based network ${\varepsilon }_{\varTheta}({\mathcal{F}}_{t_{i}}, {\mu}, {t_{i}})$ at time step $t_{i}$ to predict ${\mathcal{F}}_{t_{i-1}}$. Hence, we can get the $t_{i-1}$ state  ${\mathcal{F}}_{t_{i-1}}\gets {\mathcal{F}}_{t_{i}}-(\text{d}{\mathcal{F}}_{t_i})_{{\varepsilon }_{\varTheta}}$. The $L_{1}$ loss function is used to optimize the score-based network: 
\begin{equation}
\label{deqn_ex10a}
\mathcal{L}_{{\varepsilon }}\left( \varTheta \right) :=\sum_{i=1}^n{{\eta }_{t_i}}\mathbb{E}\left[ \lVert {\mathcal{F}}_{t_i}-\left( \text{d}{\mathcal{F}}_{t_i} \right) _{\varepsilon _{\varTheta}}-{\mathcal{F}}_{t_{i-1}}^{*} \rVert \right].
\end{equation}
Where ${{\eta }_{t_i}}>0$ is the adjustment coefficient. 

\textbf{Embeddeding Guided Conditional Network:}
As shown in Fig. \ref{The overall architecture}(b), a conditional U-Net is introduced in IR-SDE to estimate the noise ${\epsilon }_t$. Its up- and down-sampling branches are composed of alternating ResBlock and Linear attention (LA) modules, while the deepest residual convolutional layers are embedded Spatial Transformer (ST) modules. The image embedding $e_{x,y}^{I_1}$ as cross-attention guide ST Module. Specifically, we assume the input feature map of the Cross-Attention layer denotes $F_{in}\in\mathbb{R}^{(h\times w)\times c}$, $h,w$ are height and weight and $c$ is the channel number. The image embedding $e_{x,y}^{I_1}\in\mathbb{R}^{l_1}$, $l_1$ is the length of image embedding. The feature  information $F_{in}$ is transformed as $\boldsymbol{Q}\in\mathbb{R}^{(h\times w)\times c}$ and image text information as  $\boldsymbol{K}\in\mathbb{R}^{c}$, $\boldsymbol{V}\in\mathbb{R}^{c}$ through linear layers. Then, we rearrange them as $\{\boldsymbol{Q}\in\mathbb{R}^{n_h\times(h\times w)\times (c/n_h)}, \boldsymbol{K}\in\mathbb{R}^{n_h\times (c/n_h)},\boldsymbol{V}\in\mathbb{R}^{n_h\times (c/n_h)}\}$, $n_h$ is the number of head. The attention can be obtained by $\boldsymbol{Q},\boldsymbol{K}$ and $\boldsymbol{V}$, which is finally rearranged with linear layer as output $F_{out}$ that has the same dimension as $F_{in}$.

The control embedding $e_{x,y}^{I_2}$ is added with the time embedding which is encoded by Multi-Layer Perceptron (MLP), and is introduced to ResBlocks in the conditional U-Net. As shown in Fig. \ref{The overall architecture}(b), the updated time embedding $e^{t}\in\mathbb{R}^{l_2}$ and input feature map $F_{in}^{(2)}\in\mathbb{R}^{c_2\times h_2\times w_2}$ 
 are integrated with Resnet in the way as following ($l_2$ is the length of embedding, $h_2$, $w_2$are height and weight and $c_2$ is the channel number): The time embedding $e^{t}$ is rearranged and chunked as two parts denoted as $e_{1}^{t},e_{2}^{t}\in\mathbb{R}^{(l_2/4)}$, which is used for scaling and shifting $F_{in}^{(2)}$ to obtain  $F_{out}^{\left( 2 \right)}\in\mathbb{R}^{c_2\times h_2\times w_2}$.
\begin{table}
\centering
\caption{THE PERCEPTUAL INDICATOR (PI) OF DuDoCLIP}
\begin{threeparttable}  
\arrayrulecolor{black}
\arrayrulecolor{black}
\begin{tabular}{ccc}
\toprule[0.3mm]
\multirow{1}{*}{PI Type}
&\multirow{1}{*}{w/o prompt engineering}
&\multirow{1}{*}{w/ prompt engineering}
\\
\cmidrule(l){1-3}
\multirow{2}{*}{PI$_q$}
  &\multirow{2}{*}{0.48}
  &\multirow{2}{*}{0.36}
                           \\ \\
  \cmidrule(l){1-3}
\multirow{2}{*}{PI$_s$}       
  &\multirow{2}{*}{0.13}
  &\multirow{2}{*}{0.10}                              \\ \\
\bottomrule[0.3mm]
\end{tabular}
\arrayrulecolor{black}
\label{Table_add2}
 \begin{tablenotes}
        \footnotesize
        \item *PI$_q$ and PI$_s$ represent the PI indicator of DuDoCLIP on metal quantity and metal size.
      \end{tablenotes}
  \end{threeparttable}
  \vspace{-1.5em}
\end{table}
\vspace{-1em}
\subsection{OSRO Stage's Pipeline of DuDoCROP}
\subsubsection{Introduction of Raw Data Fidelity}
We devise a downstream MAR task to fully leverage the raw data fidelity and dual-domain prior information obtained from the DAPG stage. We first denote the well-trained DuDoCLIP-guided IR-SDE diffusion model as $\boldsymbol{\hat{\varepsilon}}_{\varTheta}\left( \cdot \right)$, sinogram-domain IR-SDE model as $\boldsymbol{\hat{\varepsilon}}_{\varTheta_1}\left( \cdot \right)$ with parameter $\varTheta_1$ and image-domain IR-SDE model as $\boldsymbol{\hat{\varepsilon}}_{\varTheta_2}\left( \cdot \right)$ with parameter $\varTheta_2$. Furthermore, to reduce the inconsistency between the distribution generated by the diffusion model and the raw data distribution, we introduce undamaged traces from the sinogram $Y_\text{ma}$ as data fidelity. As shown in Fig. \ref{The overall architecture}(a), We use a binary metal trace $M_s$ to extract the predicted trace of sinogram output and  $1-M_s$ to extract the fidelity data in sinogram-domain input, and the integrated term is the final result.  Hence, the sinogram-domain prior $Y_p$ incorporated data fidelity can be expressed as:
\begin{equation}
\label{deqn_ex13a}
\begin{aligned}
Y_{p}=\boldsymbol{\hat{\varepsilon}}_{\varTheta _1}\left( Y_{ma} \right) \odot M_s+Y_{ma}\odot \left( 1-M_s \right), 
\end{aligned}
\end{equation}
Where $Y_p$'s reconstructed image $X^{'}_p$ can be obtained by FBP algorithm: $X^{'}_p=FBP(Y_{p})$. Although $X^{'}_p$ contains plenty of secondary artifacts caused by the direct introduction of predicted trace in ${\hat{\varepsilon}}_{\varTheta _1}\left( Y_{ma} \right)$, however, the scale information from the raw data is preserved which is close to the ground truth. Hence, we can use this trick to normalize the image domain prior $X_{p}$ to transmit fidelity information. We first adopt the Min-Max Normalization operation as $Norm_{\delta}\left( \varDelta \right)$ ($\delta$, $\varDelta$ are different variable symbols):
\begin{equation}
\label{deqn_ex14a}
\begin{aligned}Norm_{\delta}\left(\Delta\right)&:=\frac{\Delta-\min\left(\Delta\right)}{\max\left(\Delta\right)-\min\left(\Delta\right)}\\&\times\left[\max\left(\delta\right)-\min\left(\delta\right)\right]+\min\left(\delta\right),\end{aligned}
\end{equation}
Then, the image-domain prior can be expressed as:
\begin{equation}
\label{deqn_ex15a}
\begin{aligned}
X_{p}=Norm_{X'_p}\left( \boldsymbol{\hat{\varepsilon}}_{\varTheta _2}\left( X_{ma} \right) \right),
\end{aligned}
\end{equation}
At the subsequent OSRO stage, we continue to employ this operation to propagate the fidelity information to downstream tasks, aiming to reduce the spurious structures caused by generative models.
\subsubsection{Residual Refining and Dual-Domain Prior Fusion}
At the same time, to reduce the secondary artifact in the sinogram-prior image $X'_p$, we utilize a high-quality image-domain prior to performing the residual operation with $X'_p$, where the conventional IR-SDE extracts the features of secondary artifacts in the residual domain. Then, the refined residual image $\widehat{X}_r= \boldsymbol{\varepsilon }_{\varTheta _3}(X_{p}-X'_{p})$ is fused with $X_{p}$ according to factor $k\in[0,1]$:
\begin{equation}
\label{deqn_ex16a}
\begin{aligned}
\widehat{X}_p=X_{p}-k\cdot Norm_{\left( X_{p}-X'_{p} \right)}\left( \widehat{X}_r \right), 
\end{aligned}
\end{equation}
Where $\boldsymbol{\varepsilon }_{\varTheta _3}$ is the residual-domain IR-SDE with parameter $\varTheta _3$. Similarly, We once again introduce the data fidelity from the image domain prior, and then we can obtain the updated $\widehat{X}_p$:
 \begin{equation}
\label{deqn_ex17a}
\begin{aligned}
\widehat{X}_p\gets Norm_{X_p}(\widehat{X}_p),
\end{aligned}
\end{equation}
Finally, $\widehat{X}_p$ is integrated with the reconstructed sinogram-prior image $X'_p$  with factor $\alpha\in [0,1] 
$ and $\beta\in [0,1]$ to get the final output $X_o$ of our DuDoCROP:
 \begin{equation}
\label{deqn_ex18a}
\begin{aligned}
X_o=\alpha \cdot \widehat{X}_p+\beta \cdot X'_p.
\end{aligned}
\end{equation}
In the training stage, we adopt a cascade learning scheme, where the DuDoCLIP, the image-domain IR-SDE, the sinogram-domain IR-SDE, and the residual-domain IR-SDE are trained sequentially in order. The output of the previous stage is used as input data for training the next stage. Specifically, Eq. (\ref{deqn_ex2a}) is used for DuDoCLIP training. 

\begin{algorithm}[t] 
\caption{The training pipeline of DuDoCROP.}\label{alg:alg1} \begin{algorithmic}  \STATE \hspace{-0.35cm}{\textbf{Require:}} Prepare images and corresponding sinograms $ {X}_{{\text{ma}}}$, $ {X_{ref}}$,  ${Y}_{\text{ma}}=FP({X}_{\text{ma}})$, ${Y_\text{ref}}=FP({Y}_\text{ref})$.   \STATE \hspace{-0.35cm}{\textbf{The DAPG stage:}} \STATE \hspace{-0.35cm}{\it DuDoCLIP model fine-tuning} \STATE {\textsc{1.}} Obtain $ {X}_{\text{r}}={X}_{\text{ma}}-{X_\text{ref}}$, $ {Y}_{\text{r}}={Y}_{\text{ma}}-{Y_\text{ref}}$.  \STATE {\textsc{2.}} Prepare text descriptions by LLaVA:
$\{{X}_{\text{ma}}, {Y}_{\text{ma}}\} \to T_{1(x,y)}$, $\{{X}_{\text{r}}, {Y}_{\text{r}}\} \to T_{2(x,y)}$.\STATE {\textsc{3.}} Generate text embeddings with pre-trained CLIP of text encoder: $ T_{1(x,y)} \to
\left\{ e_{x}^{T_{1}},e_{y}^{T_{1}} \right\} $, $T_{2(x,y)} \to \left\{e_{x}^{T_{2}},e_{y}^{T_{2}} \right\}$.
\STATE {\textsc{4.}} Generate image embeddings with 
\begin{itemize}
\item CLIP of image encoder: $ \{{X}_{\text{ma}}, {Y}_{\text{ma}}\} \to
\left\{ e_{x}^{I_{1}},e_{y}^{I_{1}} \right\} $, \item Control net:  $\{{X}_{\text{ma}}, {Y}_{\text{ma}}\} \to \left\{e_{x}^{I_{2}},e_{y}^{I_{2}} \right\}. 
$ \end{itemize}\STATE {\textsc{5.}} Dual-domain contrastive learning with $\mathcal{L}_{\text{con\,\,}}$ in Eq. (\ref{deqn_ex2a}). 
\STATE \hspace{-0.35cm} {\it Dual-domain IR-SDE training} \STATE {\textsc{1.}} Parameter settings in forward process: $\lambda$, $n$, etc. \STATE {\textsc{2.}} Train $\boldsymbol{\hat{\varepsilon}}_{\varTheta_1}$,$\boldsymbol{\hat{\varepsilon}}_{\varTheta_2}$ with text and image embeddings.  
\STATE \hspace{-0.35cm}{\textbf{The OSRD stage:}} 
\STATE {\textsc{1.}} Introduce data fidelity by Eq. (\ref{deqn_ex13a})-(\ref{deqn_ex15a}) to obtain $X_p$, $X'_p$.\STATE {\textsc{2.}} Prepare residual inputs $X_r^{in} = X_p-X'_p$ and references $X_r^\text{ref} = X_p-X_\text{ref}$. \STATE {\textsc{3.}} Train $\boldsymbol{{\varepsilon}}_{\varTheta_3}$ in residual domain with $\mathcal{L}_{\boldsymbol{\varepsilon }}\left( \varTheta_3 \right)$ in Eq. (\ref{deqn_ex19a}). 
\end{algorithmic}
\label{alg1}
\end{algorithm}
The loss function of dual-domain IR-SDE can be referred to as Eq.(\ref{deqn_ex10a}), but has different parameters $\varTheta_1$ and $\varTheta_2$ corresponding to sinogram domain and image domain. We fix the adjustment coefficient ${\boldsymbol{\eta }_{t_i}}$ as 1 in the experiments.  As for residual-domain IR-SDE, the residual inputs $X_r^{in} = X_p-X'_p$ are the same number as the constructed references $X_r^{ref} = X_p-X_{ref}$. Hence, the loss function of residual diffusion $\mathcal{L}_{\boldsymbol{\varepsilon }}$ can be written as:
\begin{equation}
\label{deqn_ex19a}
\!\!\!\!\mathcal{L}_{\boldsymbol{\varepsilon }}\left( \varTheta_3 \right) =\sum_{i=1}^n\mathbb{E}\left[ \lVert {X_{r(t_i)}^{in}}-\left( \text{d}{X_{r(t_i)}^{in}} \right) _{\varepsilon _{\varTheta_3}}\!\!\!\!-{X_{r(t_{i-1})}^{in(*)}} \rVert \right].
\end{equation}
Where, the expression of ${X_{r(t_{i-1})}^{in(*)}}$ is similar as Eq. (\ref{deqn_ex9a}), and the zero state ${X_{r(t_{0})}^{in(*)}}$ is so-called residual reference $X_r^{ref}$. Therefore, the total
loss function $\mathcal{L}_{total}$ of the proposed DuDoCROP is represented as follows:
\begin{equation}
\label{deqn_ex20a}
\mathcal{L}_{\text {total }}=\left\{\begin{array}{ll}
\mathcal{L}_{\text {con }}(\varphi), & \text { Stage } 1 \\
\mathcal{L}_{\hat{\varepsilon}}\left(\Theta_{1}\right)+\mathcal{L}_{\hat{\varepsilon}}\left(\Theta_{2}\right), & \text { Stage } 2 \\
\mathcal{L}_{\varepsilon}\left(\Theta_{3}\right), & \text { Stage } 3
\end{array}\right.
\end{equation}
 Stage 1: DuDoCLIP training. Stage 2: DuDoCLIP-assisted Sinogram-domain and Image-domain IR-SDE training. Stage 3: Residual domain IR-SDE training. 

 The training and testing pipeline of our DuDoCROP is shown in Algorithm \ref{alg1} and \ref{alg2}, respectively.
 \vspace{-0.5em}
\begin{algorithm}[!t] 
\caption{The testing pipeline of DuDoCROP.}\label{alg:alg1} \begin{algorithmic}  \STATE \hspace{-0.35cm}{\textbf{Require:}} Input metal-affected images and sinograms $ {X}_{\text{ma}}$, ${Y}_{\text{ma}}=FP({X}_{\text{ma}})$.   \STATE \hspace{-0.35cm}{\textbf{Parameters:}} The same network parameter settings in the training stage, and manual-adjusted factors $k$, $\alpha$, $\beta$.
 \STATE \hspace{-0.35cm}
{\textbf{Testing:}}
\STATE{\textsc{1.}} Obtain dual-dmain priors $X_p$, $Y_p$ using $\boldsymbol{\hat{\varepsilon}}_{\varTheta_1}$, $\boldsymbol{\hat{\varepsilon}}_{\varTheta_2}$ guided by DuDoCLIP. 
\STATE{\textsc{2.}} Introduce data fidelity by Eq. (\ref{deqn_ex13a})-(\ref{deqn_ex15a}) to obtain $X'_p$. \STATE{\textsc{3.}} Refine residual images $X_r^{in}$ using $\boldsymbol{{\varepsilon}}_{\varTheta_3}$ to obtain $\widehat{X}_r$.\STATE{\textsc{4.}} Get the final result $X_o$ by Eq. (\ref{deqn_ex16a})-(\ref{deqn_ex18a}). 
\end{algorithmic}
\label{alg2}
\end{algorithm}
\subsection{Perceptual Metric for Evaluating Model Generalization}
To verify the perception and generalization capabilities of our model on different morphologies of metal artifacts, we define a new perceptual metric. First, a metric of learned perceptual image patch similarity (LPIPS) is utilized to evaluate the quality and similarity of images based on their inherent features. However, it cannot quantify the sensitivity of the image to different metal implants. Based on LPIPS (Denoted LP), we design perceptual indicators PI$_q$ and PI$_s$, specifically for different metal quantities and sizes. As shown in Fig. \ref{The designing pipeline}, we select $N_s$ CT slices and perform the following operations on the $j$th slice with different metal sizes or quantities: As for the perceptual indicator of metal quantity, we adopt the method in Fig. \ref{The prompt engineering} to detect the local regions of metal implants. We assume the number of metals in slice $j$ is $N_{q}^{\left( j \right)}$. The average local indicator $
\frac{1}{N_{q}^{\left( j \right)}}\sum_i{LP_{q\left( Local \right)}^{\left( i,j \right)}}
$ is added with the global indicator $LP_{q\left( Global \right)}^{\left( j \right)}$. Then, PI$_q$ can be defined as:
\begin{equation}
\label{deqn_ex21a}
\begin{aligned}
LP_{q}^{(j)} & :=\frac{1}{N_{q}^{(j)}} \sum_{i} L P_{q(\text{Local})}^{(i, j)}+L P_{q(\text{Global})}^{(j)}, \\
P I_{q} & :=\frac{1}{N_{s}} \sum_{j} \frac{L P_{q}^{(j)}-L P_{q}^{(j-1)}}{N_{q}^{(j)}-N_{q}^{(j-1)}}.
\end{aligned}
\end{equation}
Similarly, as for the perceptual indicator of metal size, we only extract the largest metal area $A_{s}^{\left( j \right)}$ of the $j$th slice. The local and global LPIPS is denoted as $LP_{s\left(Local \right)}^{\left( j \right)}$ and $LP_{s\left(Global \right)}^{\left( j \right)}$. Hence, PI$_s$ can be obtained:
\begin{equation}
\label{deqn_ex22a}
\begin{aligned}
LP_{s}^{(j)} & :=LP_{s\left(Local \right)}^{\left( j \right)}+LP_{s\left(Global \right)}^{\left( j \right)}, \\
PI_{s} & :=\frac{1}{N_{s}} \sum_{j} \frac{LP_{s}^{(j)}-L P_{s}^{(j-1)}}{A_{s}^{(j)}-A_{s}^{(j-1)}}.
\end{aligned}
\end{equation}
The physical interpretations of PI$_q$ and PI$_s$ represent the sensitivity of the LPIPS metric to variations in the quantity and size of metal implants. Consequently, it can quantitatively assess the model's capacity to perceive and generalize across different morphologies of metal artifacts. Hence, a lower PI index signifies a superior generalization capability.


\begin{figure*}[!t]
\centering
\includegraphics[scale=0.28]{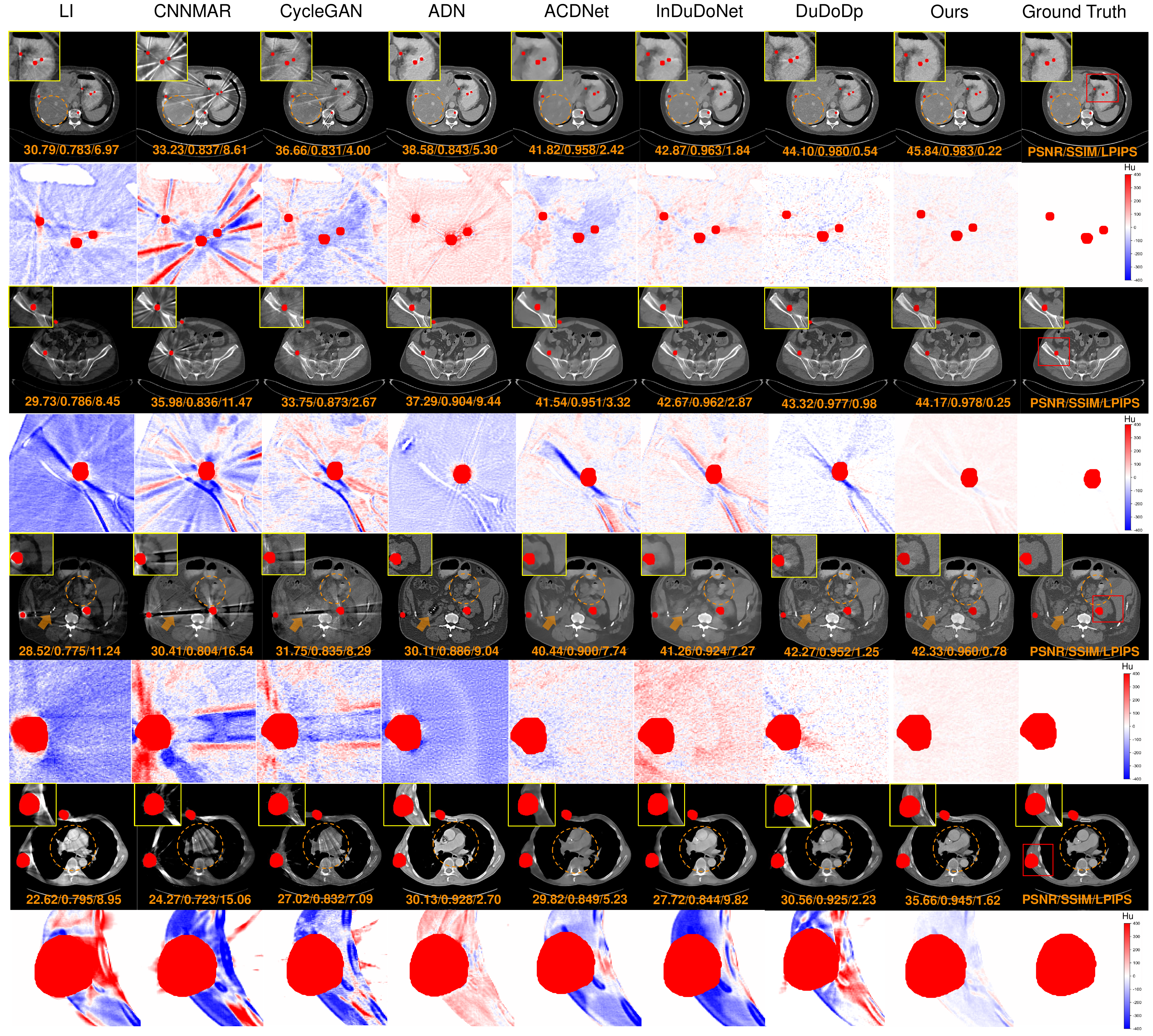}%
\caption{ Visual and numerical results of different MAR methods on AAPM dataset (D1) with different quantities of metallic implant. The display window is [-175, 275] HU. The metal masks are colored in red. Where "PSNR/SSIM/(LPIPS*100)" is demonstrated. }
\label{D1}
\vspace{-1em}
\end{figure*}
\begin{table*}
\centering
\caption{THE PERCEPTUAL INDICATOR (PI) OF DIFFERENT METHODS}
\begin{threeparttable}  
\arrayrulecolor{black}
\arrayrulecolor{black}
\begin{tabular}{ccccccccc}
\toprule[0.3mm]
\multirow{1}{*}{PI Type}
&\multirow{1}{*}{LI}
&\multirow{1}{*}{CNNMAR}
&\multirow{1}{*}{CycleGAN}
&\multirow{1}{*}{ADN}
&\multirow{1}{*}{ACDNet}
&\multirow{1}{*}{InDuDoNet}
&\multirow{1}{*}{DuDoDp}
&\multirow{1}{*}{Ours}
\\
\cmidrule(l){1-9}
\multirow{2}{*}{PI$_q$}
  &\multirow{2}{*}{0.69}
  &\multirow{2}{*}{1.61}
    &\multirow{2}{*}{1.13}
      &\multirow{2}{*}{1.35}
        &\multirow{2}{*}{0.70}
          &\multirow{2}{*}{1.99}
           &\multirow{2}{*}{0.42}
   &\multirow{2}{*}{0.35}
                           \\ \\
  \cmidrule(l){1-9}
\multirow{2}{*}{PI$_s$}       
  &\multirow{2}{*}{1.36}
  &\multirow{2}{*}{0.54}
    &\multirow{2}{*}{0.15}
      &\multirow{2}{*}{0.53}
        &\multirow{2}{*}{0.76}
          &\multirow{2}{*}{0.75}
           &\multirow{2}{*}{0.14}
   &\multirow{2}{*}{0.09} 
                              \\ \\
\bottomrule[0.3mm]
\end{tabular}
\arrayrulecolor{black}
\label{Table_add}
 \begin{tablenotes}
        \footnotesize
        \item *PI$_q$ and PI$_s$ represent the PI indicator of model on metal quantity and metal size.
      \end{tablenotes}
  \end{threeparttable}
  \vspace{-1.5em}
\end{table*}
\section{Experiments and Results}
\subsection{Data and Preprocessing}
Our experiments are conducted on a public \href{https://www.aapm.org/GrandChallenge/CT-MAR/}{AAPM Grand Challenge datasets} (CT-MAR) 
  and a clinic sub-dataset on the \href{https://github.com/MIRACLE-Center/CTPelvic1K}{CTPelvic1K} datasets. The AAPM datasets contain 13,238 512$\times$512 slices of body anatomies, including lung, abdomen, liver, and pelvis images from \href{https://nihcc.app.box.com/v/DeepLesion}{NIH DeepLesion dataset}, and 1,762 head images from the \href{https://zenodo.org/records/1199398}{UCLH Stroke EIT} Dataset. The size of the sinogram is 1000$\times$900. In Table \ref{Table1}, we uniformly sampled a variety of body datasets and a relatively small quantity of head datasets from the 14,000 official data for training. Besides them, another 642 body (body$^2$) datasets are extracted from the above official data for testing, and  864 body (body$^1$) and 136 head datasets are official testing datasets. Furthermore, we chose 200 body datasets primarily focusing on the clinical thorax from CTPelvic1K for testing, which aims to validate the generalization ability of our method. The metal-affected images, clean images and corresponding sinograms have been provided by AAPM, and there are only metal-affected images provided by CTPelvic1K. The masks can be easily segmented by threshold from metal-affected images. In IR-SDE training and testing processes, the image inputs are converted from Hounsfield Unit (HU) to attenuation coefficient (mm$^{-1}$) and normalized in the range $[0, 1]$. 
  \vspace{-1em}

  \subsection{Experimental Settings}
\subsubsection{Implementation Details}
  Our experiments are conducted on a remote server with one NVIDIA RTX A6000 GPU and 48G of memory. The proposed DuDoCROP is implemented using Python 3.8.5 and PyTorch 2.0. The preparation of image content description and feature tuple description uses the LLaVA-v1.5-13b visual-language model to achieve image-text conversion. The pre-trained text encoder and image encoder in DuDoCLIP employ a backbone of the Vision Transformer version of ViT-B-32.
    \subsubsection{Model Training and Testing}
    Our proposed method comprises three
cascaded training stages, as illustrated in Fig. \ref{The overall architecture}(a). First, the DuDoCLIP is fine-tuned on pre-trained CLIP with paired 3,003 metal-feature images and sinograms to obtain a well-trained control net. During the training phase, we set the batch size as 200 and the learning rate as 2 $\times$ 10$^{-5}$ with the maximum epoch 1000. Then, the image embedding and control embedding of the image encoder and control net are introduced to dual-domain IR-SDE diffusion model.  During the IR-SDE training phase, the same training dataset is used. A cosine noise schedule is applied with a maximum timestamp of 100, the parameter $\lambda$ is set to 50. The images are randomly cropped and rotated to a size of 256 $\times$ 256 with a batch size 16. An AdamW optimizer with an initial learning rate 2 $\times$ 10$^{-4}$ is set and a Cosine Annealing-based scheduling mechanism is used. Other parameters such as $\beta_{1}$=0.9, $\beta_{2}$=0.99 and the total iteration 70,0000 are also applied. The parameter settings for the sampling process from Gaussian noise with a variance of 25 during testing are the same as the training process. We train our model with 400 and 600 epochs on image and sinogram domains, respectively, for about 3 days. Finally, the residual-domain IR-SDE is set to the same parameters as the image-domain IR-SDE and is trained for about 200 epochs.
    \subsubsection{Performance Metrics}
    To quantify the image quality,
Peak signal noise ratio (PSNR),  structural similarity index
measure (SSIM), and learned perceptual image patch similarity (LPIPS) are utilized for quantitative evaluation. On the other hand, the proposed perceptual indicator (PI) is utilized to measure model perception and generalization ability.
\vspace{-1em}
 \subsection{Comparison Methods}
 The effectiveness of testing datasets is evaluated by comparing them utilizing a variety of competing methodologies with the proposed method. We select 7 state-of-the-art methods including a standard conventional approach based on Linear Interpolation (LI)\cite{1987Reduction}, and various DL-based approaches such as CNN\cite{8331163}, CycleGAN\cite{zhu2017unpaired}, ADN\cite{8788607}, ACDNet\cite{wang2022adaptive}, InDuDoNet\cite{inbook} and DuDoDp\cite{liu2024unsupervised}. A brief introduction is given to these models:
 \subsubsection{Conventional method}
 LI is a method that utilizes linear interpolation technology to complete the sinogram trace disrupted by metal artifacts, thereby reconstructing high-quality images. 
  \subsubsection{Unsupervised method}
In MAR, these methods do not rely on paired training data and labels. Only the sample distributions of the source domain and target domain need to be learned. i) CycleGAN achieves the conversion between domains through two generators and two discriminators. ii) ADN encodes metal-affected images into artifact and content domains in the latent space, enabling separation of the interference caused by metal implants. iii) DuDoDp is a novel score-based method proposed as a prior plugging diffusion model in dual domains. 
\subsubsection{Supervised method}
Supervised metal artifact reduction methods are currently the mainstream approach. i) CNN directly extracts features from artifact images, serving as a representative backbone for MAR. ii) ACDNet is an adaptive convolutional dictionary network that explores the prior structures of metal artifacts. iii) InDuDoNet adopts an iterative optimization method alternately updating between the image domain and sinogram domain to achieve effective artifact reduction.

Additionally, the proposed DuDoCROP introduces IR-SDE  model guided by a small amount of labeled information, while essentially still learning the statistical laws and inherent patterns to automatically restore clean images, making it more closely resemble an unsupervised approach.
\subsection{The results of prompt engineering}
As shown in Fig. \ref{The prompt engineering}, we designed prompts based on the structure of "Qualifier + Instruction" which can effectively improve the accuracy of image content description and feature description. The ratio of positive and negative words in the text descriptions will serve as our evaluation criteria. Positive words refer to those that accurately describe the organs and structures in CT images, while negative words are defined as those that incorrectly or inaccurately describe the images. We selected 1000 CT images and performed different prompts (Instruction w/ or w/o qualifier), filtering out the positive and negative words from the descriptions. The statistical results are shown in Fig. \ref{Statistical results showing}(a). After introducing qualifiers into the prompts, the accuracy of describing the structure of CT images improved by about 75.2\%. On the other hand, the separation and extraction of metal artifacts also contribute to improving the accuracy of tuple feature descriptions. As shown in Fig. \ref{Statistical results showing}(b), describing the residual map $X_r$ containing metal artifacts in a 61.5\% increase in accuracy compared to directly describing the artifact-affected images $X_\textbf{ma}$.

The accurate text description obtained through prompt engineering can facilitate the effect of metal artifact reduction. Fig. \ref{The ablation experiment} demonstrates the DuDoCLIP performance on a specific slice w/ or w/o prompt engineering. The visual effect and numerical indicators of PI in Table \ref{Table_add2} show that the precision of descriptions can enhance the model’s perception
capabilities. Especially for the incorrect anatomical structures in the yellow circle, such errors have been eliminated after prompt engineering.
\vspace{-1em}
\subsection{Simulated and Clinical Experiments}
\subsubsection{Model effect on different metal morphologies}
To fully validate the effect of DuDoCROP in the circumstance of different metal quantities and sizes,  we selected slices with 2 and 5 different metal numbers, and 4 different metal sizes. We conduct the SOTA experiments on the AAPM dataset of body$^1$(D1). Fig. \ref{D1} shows representative visual and numerical results selected from D1. Visual results indicate that LI, CNN, and CycleGAN exhibit poor perception of metal artifacts. ADN, though capable of capturing abundant details, suffers from low stability and tends to generate bizarre artifacts. ACDNet and InDuDoNet can effectively suppress artifacts but result in over-smoothed images. DuDoDp as an unsupervised method has poor generalization ability towards different artifacts. The proposed method can better reduce metal artifacts while generating relatively realistic image structures especially where large metal implants cause severe disruption. The yellow circles and arrows indicate the restored image structures. On the other hand, the numerical results show that the proposed method can also approximate pixel-level close to the ground truth since DuDoCROP introduces robust data fidelity from the raw sinogram. 

Furthermore, to more intuitively evaluate the perception and generalization performance of comparison methods, we select the same number of slices in the AAPM dataset of D1 and body$_2$ (D2) and utilize the PI index for evaluation. As shown in Table \ref{Table_add}, our method achieves the best PI$_s$ and PI$_q$ indices compared to other methods, indicating that our DuDoCROP possesses artifact perception and generalization capabilities.
Based on both visual effects and numerical metrics, it can be concluded that our method demonstrates superior performance across different quantities and sizes of metal implants.
\subsubsection{Model effect on head dataset}
\begin{figure}[!t]
\centering
\includegraphics[scale=0.20]{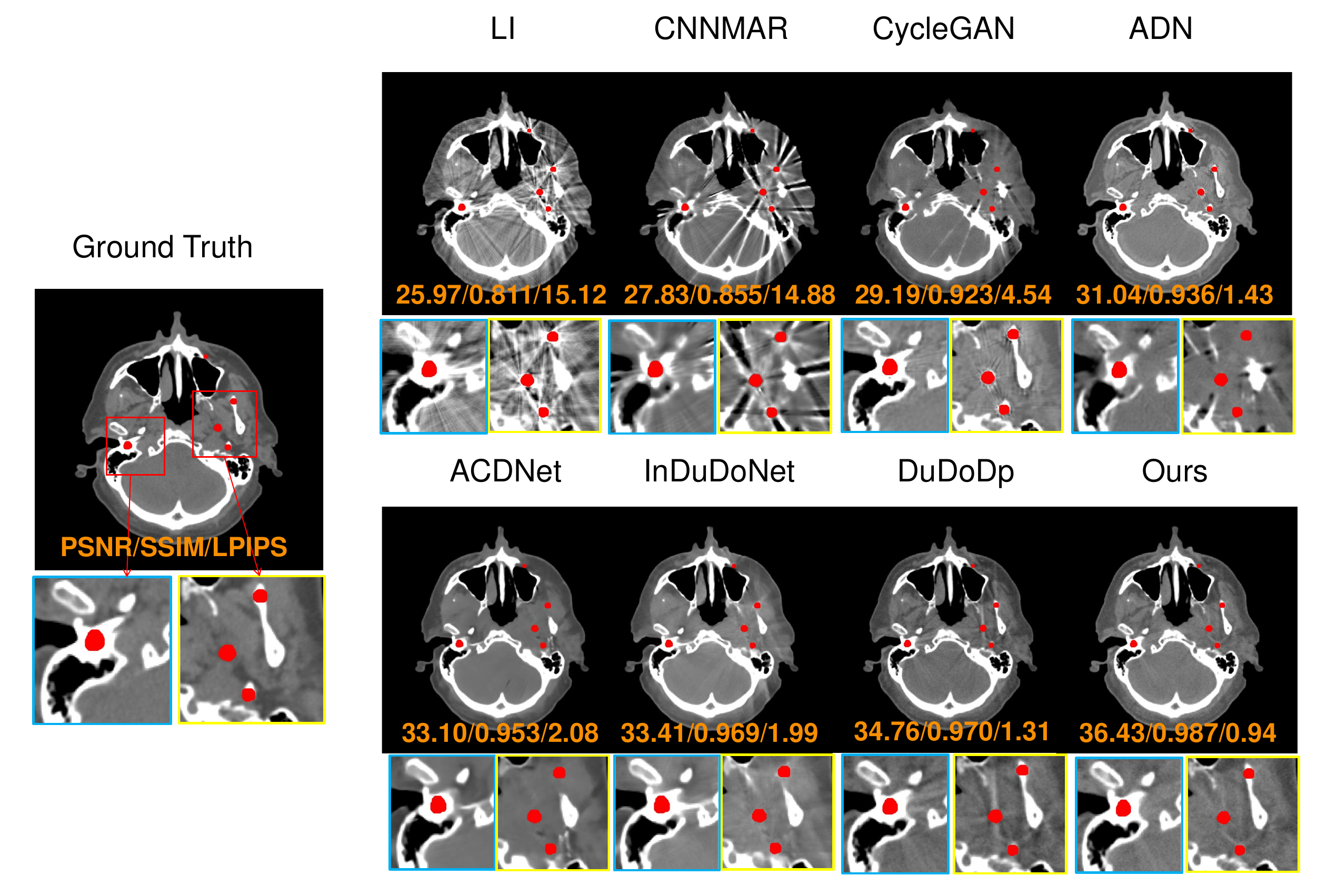}%
\vspace{-0.5em}
\caption{Visual and numerical results of different MAR methods on the head dataset (D3) with metallic implants. The display window is [-175, 275] HU. The metal masks are colored in red. Where "PSNR/SSIM/(LPIPS*100)" is demonstrated.}
\label{D3}
\vspace{-1.5em}
\end{figure}

We also conducted simulated experiments on a small amount of head data (D3) that has completely different morphological structures of the body.  Fig. \ref{D3} shows both visual effects and numerical metrics, it can be concluded that our method outperforms other algorithms. We only utilized a few hundred head images for training, demonstrating the model's remarkable ability in few-shot learning and data generalization.
Furthermore, Table \ref{Table2} presents quantitative metrics results across different datasets, which also underscore the superiority of our algorithm.
\subsubsection{Model effect on clinical dataset}
\begin{figure*}[!t]
\centering
\includegraphics[scale=0.32]{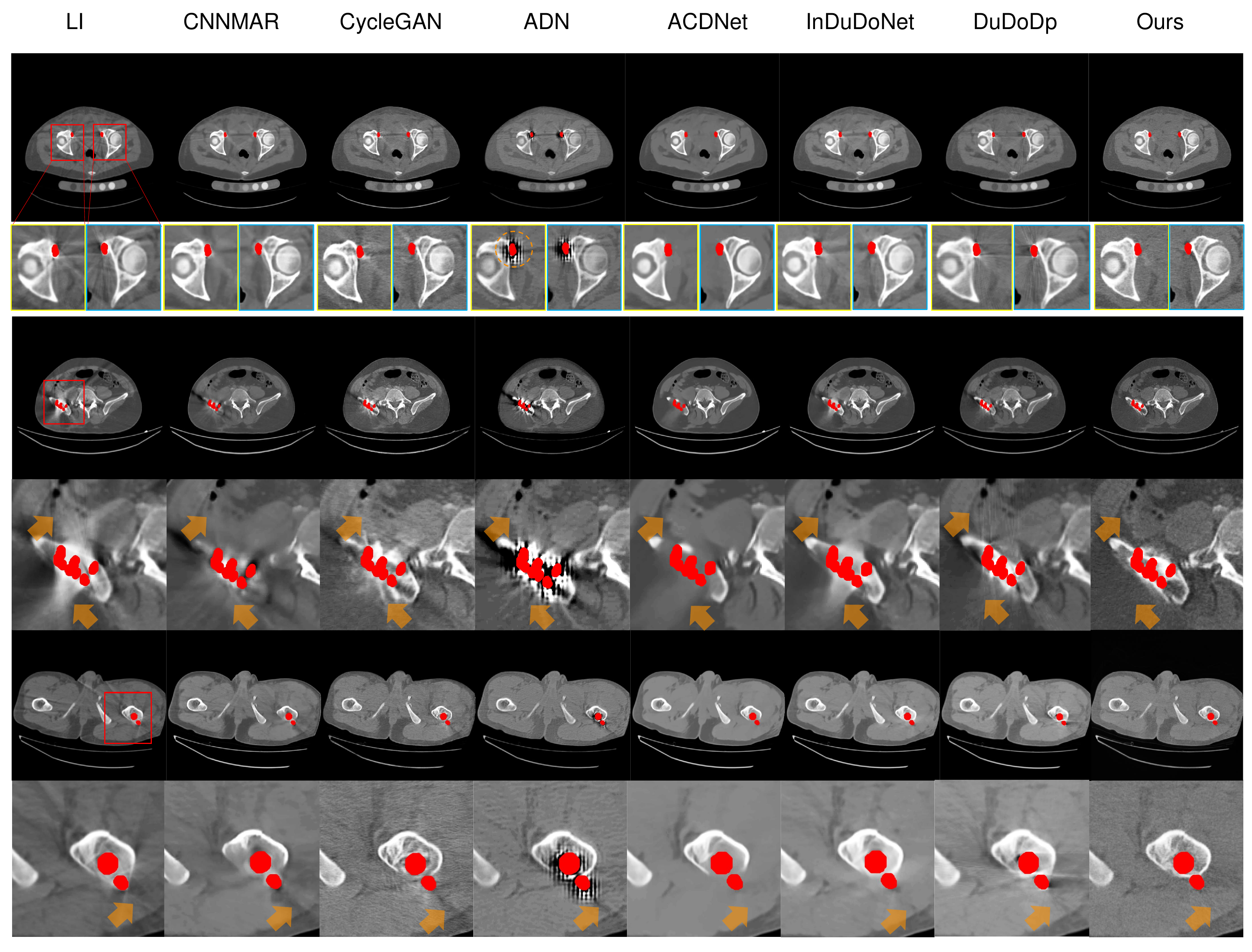}%
\vspace{-1em}
\caption{Visual results of different MAR methods on CTPelvic1K clinic dataset (D4) with metallic implants. The display window is [-175, 275] HU. The metal masks are colored in red.}
\label{D4}
\vspace{-1em}
\end{figure*}

We selected the slices with metal artifacts from the clinical CTPelvic1K dataset (D4) to test the algorithm's performance directly. Fig. \ref{D4} shows the unlabeled image visual effect. It can be seen that the results of  LI, CNN, CycleGAN, DuDoDp and InDuDoNet contain radial artifacts. The results of ADN still exhibit streak artifacts, demonstrating poor generalization capabilities. ACDNet achieves good smoothing effects but loses most structural details. In summary, our DuDoCROP demonstrates excellent visual effects and generalization capabilities, making it a promising candidate for clinical applications.

\begin{table*}
\centering
\caption{PSNR/SSIM/(LPIPS*100) IN (MEAN±SDS) FORM OF DIFFERENT MAR METHODS ON DIFFERENT DATASETS, WHERE BOLD REPRESENTS THE BEST RESULT AMONG THE DISPLAYED METHODS}
\arrayrulecolor{black}
\begin{tabular}{cccccccccc} 
\toprule[0.3mm]
\multirow{2}{*}{Dataset}                   
      &\multirow{2}{*} {Indicator}
      &\multicolumn{7}{c} {Algorithm}
      \\
      \cmidrule(l){3-10}
      &&\multicolumn{1}{c} {LI\cite{1987Reduction}}                &\multicolumn{1}{c} {CNN\cite{8331163}}                & \multicolumn{1}{c}{CycleGAN\cite{zhu2017unpaired}}                &\multicolumn{1}{c} {ADN\cite{8788607}}                 & \multicolumn{1}{c}{ACDNet\cite{8788607}}                &\multicolumn{1}{c} {InDuDoNet\cite{wang2022adaptive}}
      &\multicolumn{1}{c} {DuDoDp\cite{liu2024unsupervised}}    &\multicolumn{1}{c} {Ours}                  \\
      \cmidrule(l){1-10}
                                      
\multirow{3}{*}{body$^{1}$} &
PSNR↑                     & 28.27±3.83          & 30.23±3.90          & 30.62±4.22          & 31.49±2.85          & 41.84±3.07          & 41.88±2.30          &
40.23±4.55          &
\textbf{42.51±1.75}           \\ 
&SSIM↑                      & 0.794±0.067          & 0.832±0.075          & 0.833±0.056 &0.870±0.066          & 0.942±0.040          & 0.947±0.025          & 
0.972±0.014          & \textbf{0.980±0.008}                     \\ 
&LPIPS↓              & 7.63±2.56          & 8.09±3.60          & 3.12±1.27          & 6.83±3.85          & 2.82±2.21          & 2.44±1.48          & 
0.98±0.25          & \textbf{0.34±0.14}            \\ 
\cmidrule(l){1-10}
\multirow{3}{*}{body$^{2}$} &
PSNR↑                     & 28.14±3.99          & 30.24±3.95          & 31.18±4.00         & 31.27±2.86         & 41.37±3.47          & 41.57±2.61         & 
40.38±3.23         & \textbf{43.58±2.47}           \\ 
&SSIM↑                      & 0.786±0.075          & 0.817±0.095          & 0.840±0.066         & 0.870±0.073          & 0.935±0.049          & 0.949±0.028          &
0.975±0.016          & \textbf{0.980±0.010}           \\ 
&LPIPS↓                        & 8.07±3.25          & 8.76±3.57          & 3.18±1.51          & 6.70±4.78          & 3.00±2.76        & 2.46±1.62  & 0.90±0.32  &
\textbf{0.37±0.22}            \\ 
\cmidrule(l){1-10}
\multirow{3}{*}{head} &
PSNR↑                     & 29.79±4.29          & 29.44±3.97          & 33.92±3.18          & 32.03±2.66          & 39.50±3.73          & 41.42±3.78          &
40.73±4.48          & \textbf{44.05±4.86}           \\ 
&SSIM↑                      & 0.839±0.077          & 0.793±0.147         & 0.958±0.017          & 0.952±0.035         & 0.901±0.041          & 0.982±0.007         &
0.970±0.005         & \textbf{0.994±0.002}           \\ 
&LPIPS↓              & 8.32±4.42          & 8.37±4.78          & 2.17±1.19          & 1.29±0.58          & 1.12±0.94         & 0.83±0.45          & 1.05±0.86          &\textbf{0.37±0.36}            \\ 
\arrayrulecolor{black}
\bottomrule[0.3mm]
\end{tabular}
\label{Table2}
\end{table*}

\begin{table*}
\centering
\caption{PSNR/SSIM/(LPIPS*100) IN (MEAN±SDS) FORM AND COMPUTATIONAL TIME OF DIFFERENT ABLATION METHODS ON BODY DATASETS, WHERE BOLD REPRESENTS THE BEST RESULT AMONG THE DISPLAYED METHODS.}
\begin{tabular}{cccccc} 
\toprule[0.3mm]
\multicolumn{1}{c}{}
& \multicolumn{1}{c}{Ablation}     & \multicolumn{1}{c}{PSNR}
& \multicolumn{1}{c}{SSIM}
& \multicolumn{1}{c}{LPIPS}
& \multicolumn{1}{c}{Time(s)}
\\
\cmidrule(l){1-6}

\multirow{6}{*}{\makecell[c]{DuDoCLIP-assisted \\ Prior Generation}} &  
 Image-domain IR-SDE (w/o Normalization)           & 32.91 ± 3.21          &  0.646 ± 0.141    &  4.76 ± 0.58 & 13                       \\ 
 &Image-domain IR-SDE (w/ Normalization)            & 36.21 ± 1.32          &  0.953 ± 0.028   &  1.01 ± 0.53   & 13                     \\ 
&Sino-domain IR-SDE (w/o Fidelity)    & 23.81 ± 5.23          & 0.926 ± 0.020   & 8.49 ± 2.50   & 45                \\ 
&Sino-domain IR-SDE (w/ Fidelity)         & 39.74 ± 1.54         & 0.951 ± 0.023   & 3.90 ± 1.50    & 45                      \\ 
&Image-domain IR-SDE (w/ Normalization) + DuDoCLIP     & 36.91 ± 1.70          &  \textbf{0.977 ± 0.010}    &  \textbf{0.45 ± 0.21}                      & 17    \\ 
&Sino-domain IR-SDE (w/ Fidelity) + DuDoCLIP       & \textbf{41.36 ± 2.93}         & 0.959 ± 0.015   & 2.33 ± 1.04                  & 91   \\ 

\cmidrule(l){1-6}
\multirow{4}{*}{\makecell[c]{Downstream Task\\ Optimization} } &Dual-domain Priors Fusion       & 42.57 ± 2.04          & 0.979 ± 0.009 & 0.48 ± 0.22  & 108                         \\ 
&Residual Refining (w/o Normalization)      & 35.73 ± 1.83         & 0.968 ± 0.013  & 0.89 ± 0.38   & 125                        \\ 
&Residual Refining (w/ Normalization)    & 38.94 ± 1.80         & 0.969 ± 0.015 & 0.86 ± 0.40   & 125                          \\ 
&Residual Refining (w/ Normalization) + Dual-domain Priors Fusion           & \textbf{44.32 ± 1.52}          & \textbf{0.979 ± 0.019}   & \textbf{0.40 ± 0.21}         & 125               \\
\bottomrule[0.3mm]
\end{tabular}
\label{Table3}
\vspace{-1em}

\end{table*}
\begin{figure}[!t]
\centering
\includegraphics[scale=0.22]{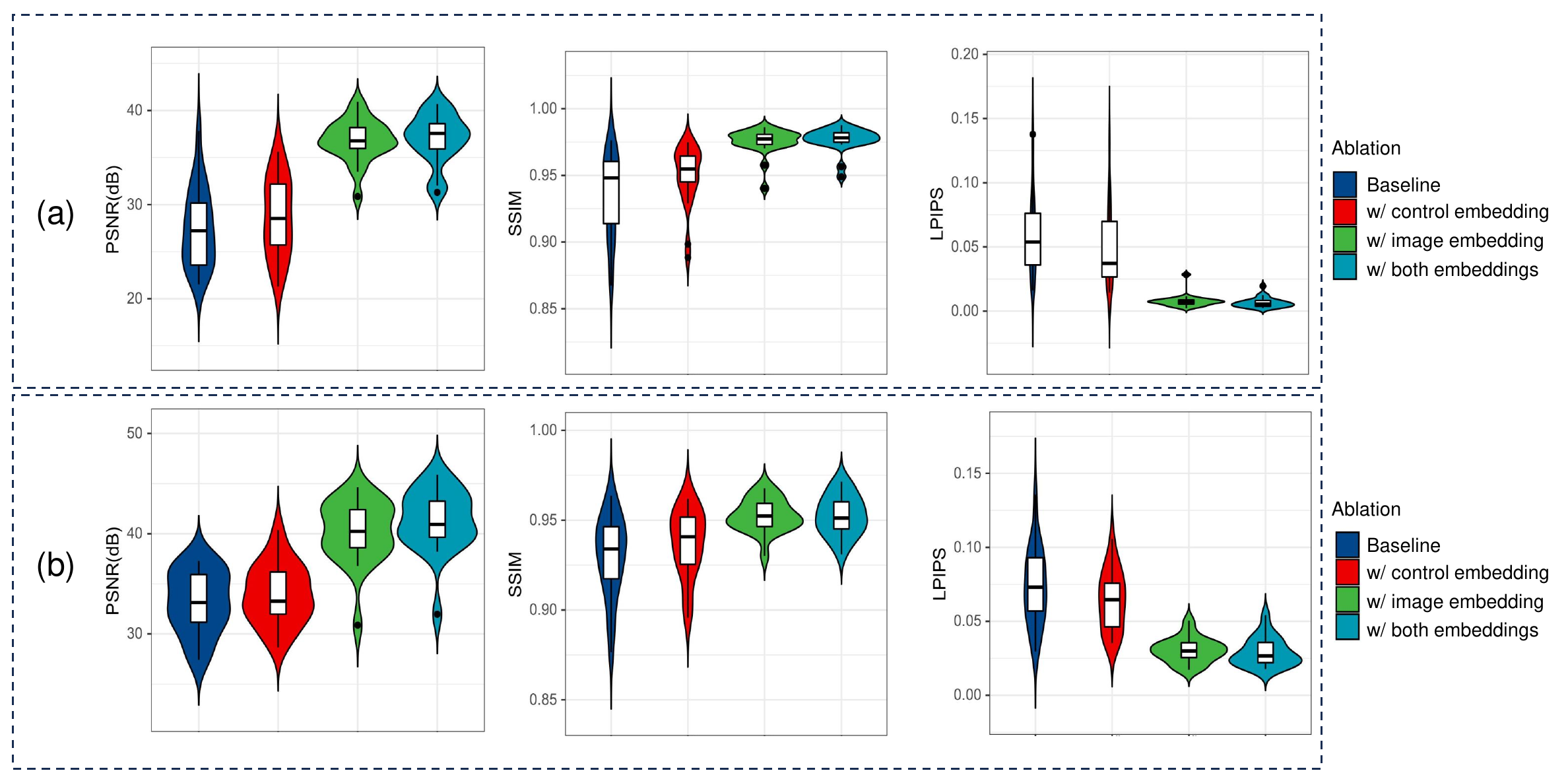}%
\vspace{-1em}
\caption{Statistical results of ablation experiments for DuDoCLIP. (a-b) are the imaging indicators of image-domain and sinogram-domain IR-SDE with the introduction of control embedding or image embedding, respectively.}
\label{Statistical results of}
\vspace{-1em}
\end{figure}
\vspace{-1em}
\subsection{Ablation Experiments}
\subsubsection{Guiding effect of DuDoCLIP}
The image embedding and control embedding of DuDoCLIP are matched with image content and metal features, respectively. We utilize these two embeddings to guide image generation within the dual-domain IR-SDE. We conducted ablation experiments on the Baseline model without embeddings, the model with image embedding, the model with control embedding, and the model with both embeddings. The statistical experiment conducted on D2 is shown in Fig. \ref{Statistical results of}(a) and (b). It can be observed that the introduction of embedding significantly contributes to model performance, especially the textual information of clean image description that notably enhances the metrics.

\subsubsection{Module ablation of DuDoCROP}
\begin{figure}[!t]
\centering
\includegraphics[scale=0.18]{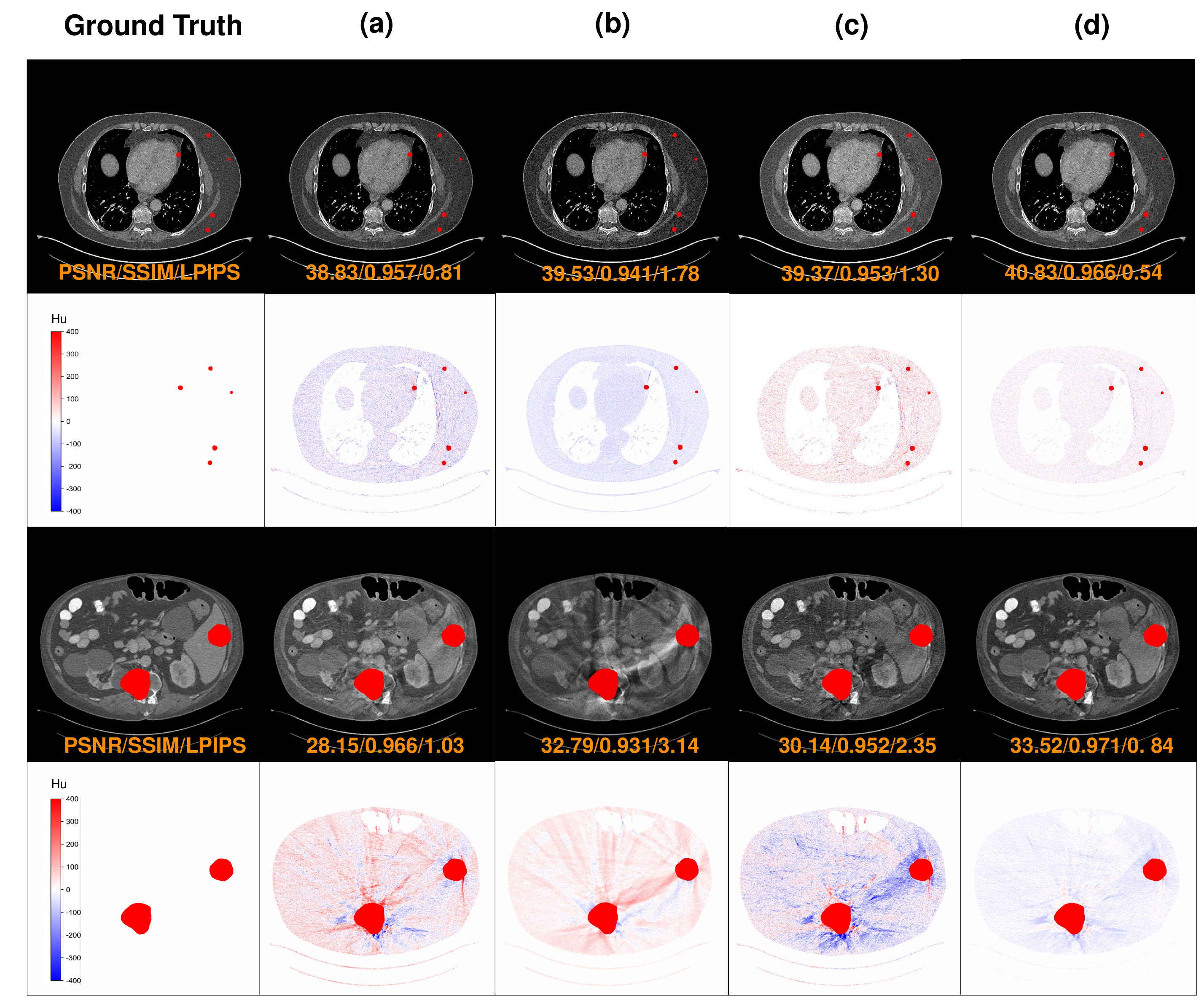}%
\vspace{-0.5em}
\caption{The visual effects of DuDoCROP using different modules. The red parts are metallic implants. (a-d) correspond to the outputs of Image-domain IR-SDE, Sino-domain IR-SDE, Residual Refining IR-SDE, and proposed DuDoCROP. The 2nd and 4th rows represent the error images. The display window of error images is [-400, 400] HU. Where "PSNR/SSIM/(LPIPS*100)" is demonstrated.}
\label{The visual effects}
\vspace{-1em}
\end{figure}

We conduct ablation studies on different modules within the overall framework of DuDoCROP. Table \ref{Table3} presents different ablation types in the stages of DAPG and OSRD, respectively (Wherein, 'w/o fidelity' refers to directly obtaining the results of IR-SDE without introducing fidelity data from the sinogram input).  It can be concluded that introducing normalization and data fidelity can improve imaging metrics. Furthermore, we find that the synthesized images obtained by combining the dual-domain fusion mechanism exhibit a PSNR improvement of at least 1-5 dB compared to single domain. The visual effects shown in Fig. \ref{The visual effects} demonstrate that with residual refining and dual-domain fusion, 
the secondary artifacts in the sinogram domain are eliminated through corrections by image priors and residual optimization. The introduction of data fidelity makes results closer to the ground truths in pixel-level distribution, with comprehensive enhancements in various metrics.
\vspace{-1em}

\section{Discussion}
Metal artifact reduction (MAR) is a typical image restoration problem in CT imaging. In clinical applications, the metal implants cause sinogram-domain measurement deviations, resulting in metal artifacts in the reconstructed image. However, due to the complexity and diversity of metal morphology, existing methods cannot perceive the artifact categories with different metal implants. Hence, they are limited in imaging effects and generalization performance. To address these issues, we innovatively leverage VLM to perceive diverse anatomical structures and metal artifacts of CT images, aligning the semantic information with image features in an embedded latent space. According to the statistics in Fig. \ref{Perception}, the average perceptual indicator (PI) of DuDoCROP is improved by at least 63.7\% compared to the IR-SDE baseline. In Table \ref{Table_add}, our method significantly addresses the issue of insufficient generalization of other SOTA methods, and demonstrates insightful imaging effects in Fig. \ref{D1}, \ref{D3} and \ref{D4}. In terms of imaging efficiency, our method requires a time cost of 125 seconds, which will provide a reference for the subsequent development of more acceleration algorithms. With the advent of more powerful VLM foundation models such as LLaVA-Next and Qwen-1.5 (with up to 110 billion parameters), we believe that DuDoCROP has the potential to demonstrate even higher capabilities in fine-grained metal artifact perception and generalization. 
\section{Conclusion}
 In this work, we primarily propose a dual-domain CLIP prior-assisted residual refining perception model named DuDoCROP, which leverages the perception ability on the morphology of anatomical structures and metal artifacts. It is a framework with DAPG and OSRD stages. DuDoCLIP is the core model in the DAPG stage fine-tuned on the image
domain and sinogram domain using contrastive learning to
extract semantic descriptions from anatomical structures and
metal artifacts. The prompt engineering and fine-grained descriptions enable DuDoCROP to perceive the artifacts more precisely. In addition, a downstream task at the OSRO stage is designed to introduce raw data fidelity and residual optimization, further enhancing the model performance and generating more realistic structures. A new perceptual indicator is designed to evaluate the perception and generalization ability. Compared to the existing state-of-the-art MAR methods, DuDoCROP exhibits superior performance in visual effects and numerical results. In the future, our DuDoCROP can serve as a paradigm for integration with the foundational model applied to various imaging tasks, such as image denoising, image reconstruction, super-resolution, {\it etc.} 
\vspace{-0.5em}


\begin{thebibliography}{10}
\providecommand{\url}[1]{#1}
\csname url@samestyle\endcsname
\providecommand{\newblock}{\relax}
\providecommand{\bibinfo}[2]{#2}
\providecommand{\BIBentrySTDinterwordspacing}{\spaceskip=0pt\relax}
\providecommand{\BIBentryALTinterwordstretchfactor}{4}
\providecommand{\BIBentryALTinterwordspacing}{\spaceskip=\fontdimen2\font plus
\BIBentryALTinterwordstretchfactor\fontdimen3\font minus
  \fontdimen4\font\relax}
\providecommand{\BIBforeignlanguage}[2]{{%
\expandafter\ifx\csname l@#1\endcsname\relax
\typeout{** WARNING: IEEEtran.bst: No hyphenation pattern has been}%
\typeout{** loaded for the language `#1'. Using the pattern for}%
\typeout{** the default language instead.}%
\else
\language=\csname l@#1\endcsname
\fi
#2}}
\providecommand{\BIBdecl}{\relax}
\BIBdecl

\bibitem{10361833}
Z.~Yang, W.~Xia, Z.~Lu, Y.~Chen, X.~Li, and Y.~Zhang, ``Hypernetwork-based
  physics-driven personalized federated learning for ct imaging,'' \emph{IEEE
  Transactions on Neural Networks and Learning Systems}, pp. 1--15, 2023.

\bibitem{9770134}
M.~Ikuta and J.~Zhang, ``A deep convolutional gated recurrent unit for ct image
  reconstruction,'' \emph{IEEE Transactions on Neural Networks and Learning
  Systems}, vol.~34, no.~12, pp. 10\,612--10\,625, 2023.

\bibitem{10121706}
J.~Zhang, Z.~Cui, C.~Jiang, S.~Guo, F.~Gao, and D.~Shen, ``Hierarchical
  organ-aware total-body standard-dose pet reconstruction from low-dose pet and
  ct images,'' \emph{IEEE Transactions on Neural Networks and Learning
  Systems}, pp. 1--13, 2023.

\bibitem{1987Reduction}
W.~A. Kalender, R.~Hebel, and J.~Ebersberger, ``Reduction of ct artifacts
  caused by metallic implants.'' \emph{Radiology}, vol. 164, no.~2, pp.
  576--577, 1987.

\bibitem{8331163}
Y.~Zhang and H.~Yu, ``Convolutional neural network based metal artifact
  reduction in x-ray computed tomography,'' \emph{IEEE Transactions on Medical
  Imaging}, vol.~37, no.~6, pp. 1370--1381, 2018.

\bibitem{yu2020deep}
L.~Yu, Z.~Zhang, X.~Li, and L.~Xing, ``Deep sinogram completion with image
  prior for metal artifact reduction in ct images,'' \emph{IEEE Transactions on
  medical imaging}, vol.~40, no.~1, pp. 228--238, 2020.

\bibitem{Lin2019DuDoNetDD}
W.-A. Lin, H.~Liao, C.~Peng, X.~Sun, J.~Zhang, J.~Luo, R.~Chellappa, and S.~K.
  Zhou, ``Dudonet: Dual domain network for ct metal artifact reduction,''
  \emph{2019 IEEE/CVF Conference on Computer Vision and Pattern Recognition
  (CVPR)}, pp. 10\,504--10\,513, 2019.

\bibitem{xie2024gan}
K.~Xie, L.~Gao, H.~Zhang, S.~Zhang, Q.~Xi, F.~Zhang, J.~Sun, T.~Lin, J.~Sui,
  and X.~Ni, ``Gan-based metal artifacts region inpainting in brain mri imaging
  with reflective registration,'' \emph{Medical Physics}, vol.~51, no.~3, pp.
  2066--2080, 2024.

\bibitem{togo2024concvae}
R.~Togo, N.~Nakagawa, T.~Ogawa, and M.~Haseyama, ``Concvae: Conceptual
  representation learning,'' \emph{IEEE Transactions on Neural Networks and
  Learning Systems}, 2024.

\bibitem{kingma2018glow}
D.~P. Kingma and P.~Dhariwal, ``Glow: Generative flow with invertible 1x1
  convolutions,'' \emph{Advances in neural information processing systems},
  vol.~31, 2018.

\bibitem{karageorgos2024denoising}
G.~M. Karageorgos, J.~Zhang, N.~Peters, W.~Xia, C.~Niu, H.~Paganetti, G.~Wang,
  and B.~De~Man, ``A denoising diffusion probabilistic model for metal artifact
  reduction in ct,'' \emph{IEEE Transactions on Medical Imaging}, 2024.

\bibitem{liu2024unsupervised}
X.~Liu, Y.~Xie, S.~Diao, S.~Tan, and X.~Liang, ``Unsupervised ct metal artifact
  reduction by plugging diffusion priors in dual domains,'' \emph{IEEE
  Transactions on Medical Imaging}, 2024.

\bibitem{dosovitskiy2020image}
A.~Dosovitskiy, L.~Beyer, A.~Kolesnikov, D.~Weissenborn, X.~Zhai,
  T.~Unterthiner, M.~Dehghani, M.~Minderer, G.~Heigold, S.~Gelly \emph{et~al.},
  ``An image is worth 16x16 words: Transformers for image recognition at
  scale,'' \emph{arXiv:2010.11929}, 2020.

\bibitem{tan2022dr}
H.~Tan, X.~Liu, B.~Yin, and X.~Li, ``Dr-gan: Distribution regularization for
  text-to-image generation,'' \emph{IEEE Transactions on Neural Networks and
  Learning Systems}, vol.~34, no.~12, pp. 10\,309--10\,323, 2022.

\bibitem{lu2019vilbert}
J.~Lu, D.~Batra, D.~Parikh, and S.~Lee, ``Vilbert: Pretraining task-agnostic
  visiolinguistic representations for vision-and-language tasks,''
  \emph{Advances in neural information processing systems}, vol.~32, 2019.

\bibitem{li2024llava}
C.~Li, C.~Wong, S.~Zhang, N.~Usuyama, H.~Liu, J.~Yang, T.~Naumann, H.~Poon, and
  J.~Gao, ``Llava-med: Training a large language-and-vision assistant for
  biomedicine in one day,'' \emph{Advances in Neural Information Processing
  Systems}, vol.~36, 2024.

\bibitem{wu2023visual}
C.~Wu, S.~Yin, W.~Qi, X.~Wang, Z.~Tang, and N.~Duan, ``Visual chatgpt: Talking,
  drawing and editing with visual foundation models,'' \emph{arXiv:2303.04671},
  2023.

\bibitem{liu2021medical}
G.~Liu, Y.~Liao, F.~Wang, B.~Zhang, L.~Zhang, X.~Liang, X.~Wan, S.~Li, Z.~Li,
  S.~Zhang \emph{et~al.}, ``Medical-vlbert: Medical visual language bert for
  covid-19 ct report generation with alternate learning,'' \emph{IEEE
  Transactions on Neural Networks and Learning Systems}, vol.~32, no.~9, pp.
  3786--3797, 2021.

\bibitem{zhong2023chatradio}
T.~Zhong, W.~Zhao, Y.~Zhang, Y.~Pan, P.~Dong, Z.~Jiang, X.~Kui, Y.~Shang,
  L.~Yang, Y.~Wei \emph{et~al.}, ``Chatradio-valuer: a chat large language
  model for generalizable radiology report generation based on
  multi-institution and multi-system data,'' \emph{arXiv:2310.05242}, 2023.

\bibitem{chen2024low}
Z.~Chen, T.~Chen, C.~Wang, C.~Niu, G.~Wang, and H.~Shan, ``Low-dose ct
  denoising with language-engaged dual-space alignment,''
  \emph{arXiv:2403.06128}, 2024.

\bibitem{hafner2021clip}
M.~Hafner, M.~Katsantoni, T.~K{\"o}ster, J.~Marks, J.~Mukherjee, D.~Staiger,
  J.~Ule, and M.~Zavolan, ``Clip and complementary methods,'' \emph{Nature
  Reviews Methods Primers}, vol.~1, no.~1, pp. 1--23, 2021.

\bibitem{gu2006method}
J.-w. Gu, L.~Zhang, Z.-q. Chen, Y.-x. Xing, and Z.-f. Huang, ``A method based
  on interpolation for metal artifacts reduction in ct images,'' \emph{Journal
  of x-ray Science and Technology}, vol.~14, no.~1, pp. 11--19, 2006.

\bibitem{zhang2016iterative}
H.~Zhang, L.~Wang, L.~Li, A.~Cai, G.~Hu, and B.~Yan, ``Iterative metal artifact
  reduction for x-ray computed tomography using unmatched
  projector/backprojector pairs,'' \emph{Medical physics}, vol.~43, no. 6Part1,
  pp. 3019--3033, 2016.

\bibitem{winklhofer2018combining}
S.~Winklhofer, R.~Hinzpeter, D.~Stocker, G.~Baltsavias, L.~Michels, J.-K.
  Burkhardt, L.~Regli, A.~Valavanis, and H.~Alkadhi, ``Combining monoenergetic
  extrapolations from dual-energy ct with iterative reconstructions: reduction
  of coil and clip artifacts from intracranial aneurysm therapy,''
  \emph{Neuroradiology}, vol.~60, pp. 281--291, 2018.

\bibitem{meyer2010normalized}
E.~Meyer, R.~Raupach, M.~Lell, B.~Schmidt, and M.~Kachelrie{\ss}, ``Normalized
  metal artifact reduction (nmar) in computed tomography,'' \emph{Medical
  physics}, vol.~37, no.~10, pp. 5482--5493, 2010.

\bibitem{meyer2012frequency}
E.~\vspace{0mm} Meyer, R.~Raupach, M.~Lell, B.~Schmidt, and M.~Kachelrie{\ss},
  ``Frequency split metal artifact reduction (fsmar) in computed tomography,''
  \emph{Medical physics}, vol.~39, no.~4, pp. 1904--1916, 2012.

\bibitem{lyu2023pds}
T.~Lyu, Z.~Wu, G.~Ma, C.~Jiang, X.~Zhong, Y.~Xi, Y.~Chen, and W.~Zhu,
  ``Pds-mar: a fine-grained projection-domain segmentation-based metal artifact
  reduction method for intraoperative cbct images with guidewires,''
  \emph{Physics in Medicine \& Biology}, vol.~68, no.~21, p. 215007, 2023.

\bibitem{8788607}
H.~Liao, W.-A. Lin, S.~K. Zhou, and J.~Luo, ``Adn: Artifact disentanglement
  network for unsupervised metal artifact reduction,'' \emph{IEEE Transactions
  on Medical Imaging}, vol.~39, no.~3, pp. 634--643, 2020.

\bibitem{lee2021unsupervised}
J.~Lee, J.~Gu, and J.~C. Ye, ``Unsupervised ct metal artifact learning using
  attention-guided $\beta$-cyclegan,'' \emph{IEEE Transactions on Medical
  Imaging}, vol.~40, no.~12, pp. 3932--3944, 2021.

\bibitem{song2020score}
Y.~Song, J.~Sohl-Dickstein, D.~P. Kingma, A.~Kumar, S.~Ermon, and B.~Poole,
  ``Score-based generative modeling through stochastic differential
  equations,'' \emph{arXiv:2011.13456}, 2020.

\bibitem{song2021denoising}
J.~Song, C.~Meng, and S.~Ermon, ``Denoising diffusion implicit models,'' in
  \emph{International Conference on Learning Representations}, 2021.

\bibitem{dou2024diffusion}
Z.~Dou and Y.~Song, ``Diffusion posterior sampling for linear inverse problem
  solving: A filtering perspective,'' in \emph{The Twelfth International
  Conference on Learning Representations}, 2024.

\bibitem{tong2023data}
Z.~Tong, Z.~Wu, Y.~Yang, W.~Mao, S.~Wang, Y.~Li, and Y.~Chen, ``Data-consistent
  unsupervised diffusion model for metal artifact reduction,'' in \emph{2023
  IEEE International Conference on Bioinformatics and Biomedicine
  (BIBM)}.\hskip 1em plus 0.5em minus 0.4em\relax IEEE, 2023, pp. 1467--1472.

\bibitem{choi2024dual}
Y.~Choi, D.~Kwon, and S.~J. Baek, ``Dual domain diffusion guidance for 3d cbct
  metal artifact reduction,'' in \emph{Proceedings of the IEEE/CVF Winter
  Conference on Applications of Computer Vision}, 2024, pp. 7965--7974.

\bibitem{inbook}
H.~Wang, Y.~Li, H.~Zhang, J.~Chen, K.~Ma, D.~Meng, and Y.~Zheng, ``Indudonet:
  An interpretable dual domain network for ct metal artifact reduction,'' in
  \emph{Medical Image Computing and Computer Assisted Intervention – MICCAI
  2021}, 09 2021, pp. 107--118.

\bibitem{ZHOU2022102289}
B.~Zhou, X.~Chen, S.~K. Zhou, J.~S. Duncan, and C.~Liu, ``Dudodr-net:
  Dual-domain data consistent recurrent network for simultaneous sparse view
  and metal artifact reduction in computed tomography,'' \emph{Medical Image
  Analysis}, vol.~75, p. 102289, 2022.

\bibitem{li2024quad}
Z.~Li, Q.~Gao, Y.~Wu, C.~Niu, J.~Zhang, M.~Wang, G.~Wang, and H.~Shan,
  ``Quad-net: Quad-domain network for ct metal artifact reduction,'' \emph{IEEE
  Transactions on Medical Imaging}, 2024.

\bibitem{9765584}
T.~Wang, Z.~Lu, Z.~Yang, W.~Xia, M.~Hou, H.~Sun, Y.~Liu, H.~Chen, J.~Zhou, and
  Y.~Zhang, ``Idol-net: An interactive dual-domain parallel network for ct
  metal artifact reduction,'' \emph{IEEE Transactions on Radiation and Plasma
  Medical Sciences}, vol.~6, no.~8, pp. 874--885, 2022.

\bibitem{long2023fine}
Y.~Long, J.~Han, R.~Huang, H.~Xu, Y.~Zhu, C.~Xu, and X.~Liang, ``Fine-grained
  visual--text prompt-driven self-training for open-vocabulary object
  detection,'' \emph{IEEE Transactions on Neural Networks and Learning
  Systems}, 2023.

\bibitem{rombach2022high}
R.~Rombach, A.~Blattmann, D.~Lorenz, P.~Esser, and B.~Ommer, ``High-resolution
  image synthesis with latent diffusion models,'' in \emph{Proceedings of the
  IEEE/CVF conference on computer vision and pattern recognition}, 2022, pp.
  10\,684--10\,695.

\bibitem{yu2024scaling}
F.~Yu, J.~Gu, Z.~Li, J.~Hu, X.~Kong, X.~Wang, J.~He, Y.~Qiao, and C.~Dong,
  ``Scaling up to excellence: Practicing model scaling for photo-realistic
  image restoration in the wild,'' in \emph{Proceedings of the IEEE/CVF
  Conference on Computer Vision and Pattern Recognition}, 2024, pp.
  25\,669--25\,680.

\bibitem{luo2023controlling}
Z.~\vspace{0mm} Luo, F.~K. Gustafsson, Z.~Zhao, J.~Sj{\"o}lund, and T.~B.
  Sch{\"o}n, ``Controlling vision-language models for universal image
  restoration,'' \emph{arXiv:2310.01018}, 2023.

\bibitem{zhang2023adding}
L.~Zhang, A.~Rao, and M.~Agrawala, ``Adding conditional control to
  text-to-image diffusion models,'' in \emph{Proceedings of the IEEE/CVF
  International Conference on Computer Vision}, 2023, pp. 3836--3847.

\bibitem{luo2023image}
Z.~Luo, F.~K. Gustafsson, Z.~Zhao, J.~Sj{\"o}lund, and T.~B. Sch{\"o}n, ``Image
  restoration with mean-reverting stochastic differential equations,''
  \emph{arXiv:2301.11699}, 2023.

\bibitem{anderson1982reverse}
B.~D. Anderson, ``Reverse-time diffusion equation models,'' \emph{Stochastic
  Processes and their Applications}, vol.~12, no.~3, pp. 313--326, 1982.

\bibitem{ho2020denoising}
J.~Ho, A.~Jain, and P.~Abbeel, ``Denoising diffusion probabilistic models,''
  \emph{Advances in neural information processing systems}, vol.~33, pp.
  6840--6851, 2020.

\bibitem{zhu2017unpaired}
J.-Y. Zhu, T.~Park, P.~Isola, and A.~A. Efros, ``Unpaired image-to-image
  translation using cycle-consistent adversarial networks,'' in
  \emph{Proceedings of the IEEE international conference on computer vision},
  2017, pp. 2223--2232.

\bibitem{wang2022adaptive}
H.~Wang, Y.~Li, D.~Meng, and Y.~Zheng, ``Adaptive convolutional dictionary
  network for ct metal artifact reduction,'' \emph{arXiv:2205.07471}, 2022.

\end{thebibliography}

\end{document}